%% file: main.tex
\documentclass{article}

\usepackage{style}
\usepackage{floatrow}
\newfloatcommand{capbtabbox}{table}[][\FBwidth]
\usepackage[final]{neurips_2024}

\usepackage[utf8]{inputenc} 
\usepackage[T1]{fontenc}    
\usepackage{hyperref}       
\usepackage{url}            
\usepackage{booktabs}       
\usepackage{amsfonts}       
\usepackage{nicefrac}       
\usepackage{microtype}      
\usepackage{xcolor}         
\usepackage{wrapfig}
\usepackage{bbm}
\newcommand{\methodname}{{{\textbf{ProvNeRF}}}\xspace}
\definecolor{georgecolor}{RGB}{255, 87, 51}

\definecolor{mkcolor}{RGB}{255,0, 128}

\definecolor{kecolor}{RGB}{0,128,0}

\definecolor{leocolor}{RGB}{0,0,255}

\title{ProvNeRF: Modeling per Point Provenance in NeRFs as a Stochastic Field}

\author{%
    \textbf{Kiyohiro Nakayama}$^1$~~~~~~\textbf{Mikaela Angelina Uy}$^{1, 2}$~~~~~~\textbf{Yang You}$^1$~~~~~~\textbf{Ke Li}$^3$~~~~~~\textbf{Leonida J. Guibas}$^1$\\
    $\phantom{}^1$ Stanford University~~~~~~~$\phantom{}^2$ Nvidia~~~~~~~$\phantom{}^3$ Simon Fraser University\\
    \texttt{w4756677@stanford.edu}~~~~~~~\texttt{keli@sfu.ca} \\\texttt{\{mikacuy, yangyou, guibas\}@cs.stanford.edu}
}

\begin{document}
\maketitle
\input{sections/0_abstract}
\input{sections/1_intro-mika}
\begin{figure}[t]
\begin{center}
\includegraphics[width=\textwidth]{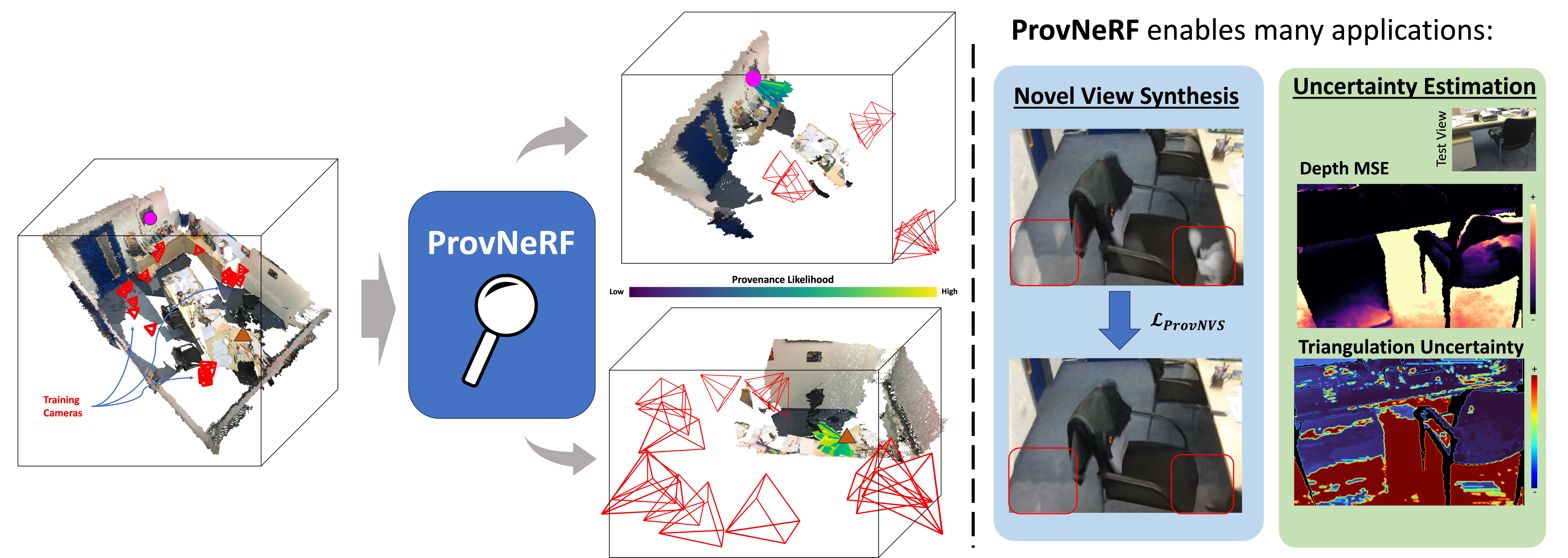}
    \caption{(\textbf{Left}) \methodname models a provenance field that outputs \emph{provenances} for each 3D point as likely samples (arrows). For 3D points (brown triangle and blue circle), the corresponding provenances (illustrated by the arrows), are locations that likely observe them. (\textbf{Right}) \methodname enables better novel view synthesis and estimating the uncertainty of the capturing process because it models the locations of likely observations that is critical for NeRF's optimization. \vspace{-0.5cm}}
    \label{fig_teaser}
\end{center}
\end{figure}
\input{sections/2_related_works}
\input{sections/3_background}
\input{sections/4_method}
\input{sections/5_results}

\input{sections/6_conclusion}
\input{sections/7_acknowledgement}

{\small
\bibliographystyle{ieee_fullname}
\bibliography{main}
}


\appendix

\input{sections/X_suppl}

\end{document}

%% file: sections/0_abstract.tex
\begin{abstract}
Neural radiance fields (NeRFs) have gained popularity with multiple works showing promising results across various applications. However, to the best of our knowledge, existing works do not explicitly model the distribution of training camera poses, or consequently the triangulation quality, a key factor affecting reconstruction quality dating back to classical vision literature.
We close this gap with ProvNeRF, an approach that models the \textbf{provenance} for each point -- i.e., the locations where it is likely visible -- of NeRFs as a stochastic field. We achieve this by extending implicit maximum likelihood estimation (IMLE) to functional space with an optimizable objective. 
We show that modeling per-point provenance during the NeRF optimization enriches the model with information on triangulation leading to improvements in novel view synthesis and uncertainty estimation under the challenging sparse, unconstrained view setting against competitive baselines.
\end{abstract}

%% file: sections/1_intro-mika.tex
\section{Introduction}
\label{sec:intro}

Neural radiance fields (NeRFs)~\cite{mildenhall2020nerf}, allowing for learning 3D scenes given only 2D images, have grown in popularity in recent years. It has shown promise in many different applications such as novel view synthesis~\cite{barron2021mipnerf, barron2023zipnerf}, depth estimation~\cite{deng2022depthsupervised}, robotics~\cite{IchnowskiAvigal2021DexNeRF, nerf-nav}, localization~\cite{liu2023nerfloc, maggio2022locnerf}, etc. Existing literature~\cite{charatan23pixelsplat,du2023cross, pan2022activenerf} show that the quality of NeRF reconstruction is correlated with the selection of training camera poses. Similar correlations are observed in the classical literature too, triangulation is highly dependent on camera poses~\cite{Praveen19, Okutomi_multiple_baseline_stereo, atanassovCamBaseline}, which greatly influences the reconstruction quality.
One common and important setting in computer vision literature~\cite{5459148, 9879641, Matsuki2021CodeMappingRD, 7898369} is the \textbf{sparse view}~\cite{8637125} setting in \textbf{unconstrained}~\cite{10.1145/1141911.1141964} environments, and triangulation is even more critical, affecting the reconstruction quality as limited input views make the system more sensitive to noise. 

Despite the correlation between triangulation and reconstruction quality, to the best of our knowledge, existing works do not explicitly model the former when optimizing the latter. In this work, we address this gap in the literature by modeling for each point the \emph{locations where it is likely visible}. We dub this as the \emph{provenances} of a point. Modeling and learning per-point provenance can help NeRF understand how the training cameras are distributed in space, which inherently links it to triangulation and reconstruction quality.


However, determining the provenances of a point $\bm{x}$ without the underlying geometry is not straightforward as many factors influence the visibility of each point in the reconstructed geometry. For example, the literature on stereo matching~\cite{Okutomi_multiple_baseline_stereo, Praveen19} has extensively studied the influences of camera locations on 3D reconstruction. One such well-known challenge arises when selecting the baseline of a pair of cameras in a stereo system. As shown in Fig.~\ref{fig:tradeoff}, points' visibility can suffer from different sets of errors when the length of the camera pair's baseline changes. For NeRFs, the dependence becomes more complex as multiple cameras' visibility needs to be estimated. To overcome this challenge, we propose to model the provenance as the samples from a \textit{probability distribution}, where a location $\bm{y}$ is assigned with a large likelihood if and only if $\bm{x}$ is likely to be visible from $\bm{y}$.

To handle the potential complexity of this distribution, we represent the provenance of $\bm{x}$ as a set of location \emph{samples}, generated from a learned probability distribution. This is distinct from the existing ``attribute" prediction extensions of NeRFs~\cite{Zhi:etal:ICCV2021,kobayashi2022distilledfeaturefields, cen2023segment} since provenance is a \emph{distribution} for every 3D point in space. Thus, this amounts to modeling an infinite collection of distributions (per-point's provenance) over all 3D points, which is mathematically, a \emph{stochastic field} over $\R^3$. In our work, we extend implicit maximum likelihood estimation (IMLE)~\cite{li2018implicit}, a sample-based generative model, to model stochastic fields by adapting the objective to functional space. Furthermore, we derive an equivalent pointwise objective that can be efficiently optimized with gradient descent and use it to model the provenance field.


We dub our method \methodname which models per-point provenance during the training stage of NeRF (Fig.~\ref{fig_teaser}). This enriches the model with information on triangulation quality when the model parameters are optimized. Once the provenance stochastic field is trained, we show that we can use it to improve novel view synthesis (Sec.~\ref{sec_nvs}) and estimate triangulation uncertainty in the capturing process (Sec.~\ref{sec_uncertainty}) under the challenging sparse, unconstrained view setting. 




%% file: sections/2_related_works.tex
\section{Related Works}
\label{sec:related_works}
\paragraph{NeRFs and their Extensions.} Neural radiance fields (NeRFs)~\cite{mildenhall2020nerf} have revolutionized the field of 3D reconstruction~\cite{hartley_zisserman_2004} and novel view synthesis~\cite{609462, space_carving} with its powerful representation of a scene using weights of an MLP that is rendered by volume rendering~\cite{max1995optical, uy-plnerf-neurips23}. 
Follow-ups on NeRF further tackle novel view synthesis under more difficult scenarios such as unconstrained photo collections~\cite{martinbrualla2020nerfw}, unbounded~\cite{barron2022mip360}, dynamic~\cite{li2023dynibar} and deformable~\cite{park2021nerfies} scenes, and reflective objects~\cite{verbin2022refnerf, bi2020neural}. Going beyond novel view synthesis, the NeRF representation has also shown great promise in different applications such as autonomous driving~\cite{tancik2022block, xiangli2022bungeenerf}, robotics~\cite{nerf-nav, IchnowskiAvigal2021DexNeRF} and editing~\cite{nerfediting, cagenerf}. Recent works have also extended NeRFs to model other fields in addition to color and opacity such as semantics~\cite{Zhi:etal:ICCV2021, zhang2022nerfusion}, normals~\cite{yu2022monosdf}, CLIP embeddings~\cite{lerf2023}, image features~\cite{kobayashi2022distilledfeaturefields} and scene flow~\cite{li2021neural}. Most of these works learn an additional function that predicts an auxiliary \emph{deterministic} output at each point that is either a scalar or a vector, trained with extra supervision using volume rendering. 
All of the above works use a deterministic field to output the additional information. However, because each point's provenance is a probabilistic distribution, we need to model a stochastic field instead of a deterministic field for provenance. 



\paragraph{Sparse View Novel View Synthesis.} NeRFs with rendering supervision alone struggle with sparse view input due to insufficient constraints in volume rendering. Several approaches have been proposed to train NeRFs under the sparse-view regime with regularization losses~\cite{Niemeyer2021Regnerf, Yang2023FreeNeRF}, semantic consistency~\cite{Jain_2021_ICCV}, and image~\cite{wang2021ibrnet} or cost volume~\cite{mvsnerf, wei2021nerfingmvs, chang2022rcmvsnet} feature constraints. Other works also constrain the optimization using priors from data~\cite{jang2021codenerf, yu2021pixelnerf} or depth~\cite{roessle2022depthpriorsnerf, scade, song2023d}. Despite addressing the setting with limited number of input views, many works are not specifically designed to tackle our desired sparse, unconstrained views setting as they either focus on object-level~\cite{jang2021codenerf, yu2021pixelnerf, Niemeyer2021Regnerf, Jain_2021_ICCV, huang2023sc}, limited camera baselines~\cite{mvsnerf, wei2021nerfingmvs}, or forward-facing scenes~\cite{wang2021ibrnet} scenes. Recent works~\cite{roessle2022depthpriorsnerf, scade, song2023d} have looked into improving the NeRF quality on a more difficult setting of sparse, unconstrained (outward-facing) input views by incorporating depth priors. However, none of these works consider the locations and orientations of training cameras in their optimization process despite it being one of the major factors influencing the NeRF's optimization in a sparse setup. 

\begin{figure}[t]
    \includegraphics[height=4cm]{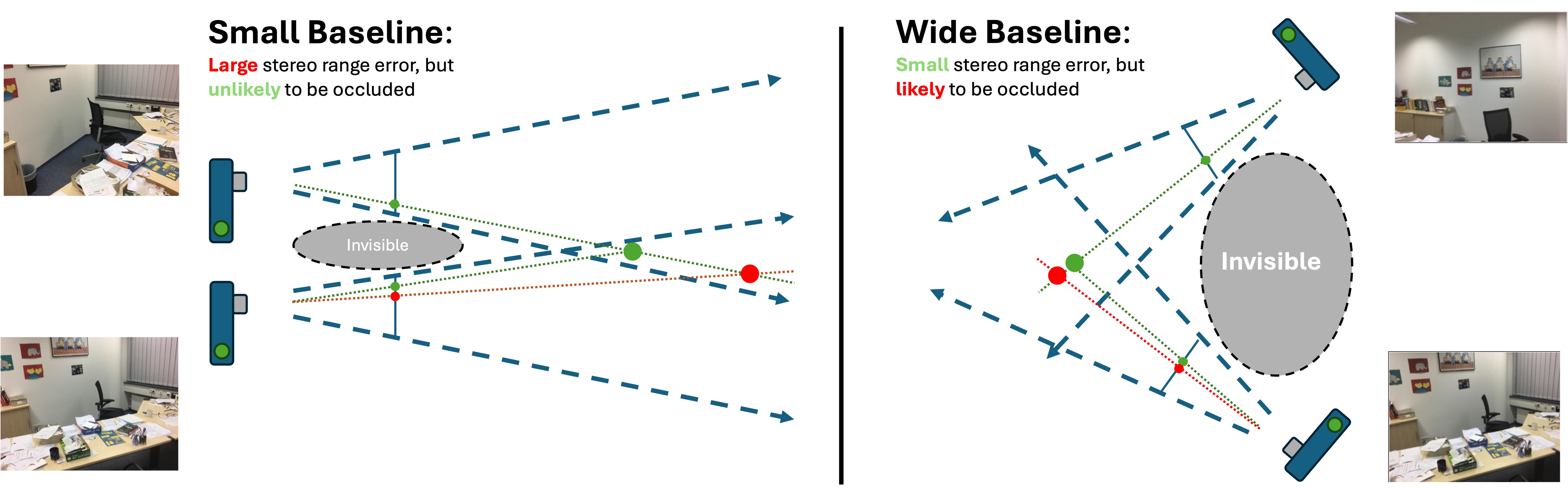}
    \caption{\textbf{Complex influence of camera baseline distance on the 3D reconstruction.} \textit{Right:} With a wide baseline, the reconstruction is more robust against 2D measurement noises. However, it is more likely to omit hidden surfaces because the invisible region is larger than a small baseline camera pair. \textit{Left:} With a small baseline, the 3D reconstruction is less likely to suffer from occlusions as the invisible region between cameras is small. However, the reconstruction can be noisy due to large stereo range errors (large deviation in depth with a small amount of noise in the 2D measurement). \vspace{-0.5cm}
    }
    \label{fig:tradeoff}
\end{figure}
\paragraph{Uncertainty Modeling in Neural Radiance Fields.} The current literature separates NeRF's uncertainty into aleatoric -- e.g., transient objects or changes in lighting -- and epistemic -- data limitation due to weak texture or limited camera views -- uncertainties. Some works~\cite{martinbrualla2020nerfw, jin2023neunbv} model aleatoric uncertainty by directly predicting the uncertainty values through a neural network. However, their approach requires training on large-scale data and is not suited for estimating the uncertainty of a specific scene. On the other hand, several works explore epistemic uncertainty estimation in NeRFs through variational inference~\cite{s-nerf, CF-NeRF, Ran2022NeurARNU}, ensemble learning~\cite{sünderhauf2022densityaware, KD23}, and Bayesian inference~\cite{bayesrays, jiang2023fisherrf, pan2022activenerf}. 
While these works estimate epistemic uncertainty, they still entangle different sources of uncertainty such as texture, camera poses, and model bias, resulting in unclear and inconsistent definitions of the uncertainty quantified. In our work, we specifically model the uncertainty caused by the capturing process that is useful in various downstream tasks~\cite{pan2022activenerf, jin2023neunbv, Ran2022NeurARNU, KD23}. 

%% file: sections/3_background.tex
\section{Preliminaries}
\label{sec:background}
\subsection{Neural Radiance Fields (NeRF)}
A neural radiance field (NeRF) is a coordinate-based neural network that learns a field in 3D space, where each point $\bm{x} \in \mathbb{R}^3$ is of certain \emph{opacity} and \emph{color}. Mathematically, a NeRF is parameterized by two functions representing the two fields $\bm{F}_{\phi, \psi} = (\bm{\sigma}_{\psi}(\bm{x}), \bm{c}_{\phi}(\bm{x}, \bm{d}))$, one for opacity $\bm{\sigma}_{\psi}:\mathbb{R}^3 \rightarrow \mathbb{R}_+$ and one for color $\bm{c}_{\phi}:\mathbb{R}^3 \times \mathbb{S}^2 \rightarrow [0,1]^3$, where $\bm{d}\in \mathbb{S}^2$ is the direction from where $\bm{x}$ is viewed from. One of the key underpinnings of NeRFs is volume rendering 
allowing for end-to-end differentiable learning with only training images. Concretely, given a set of $M$ images $I_1, I_2, ..., I_M$ and their corresponding camera poses $P_1, P_2, ..., P_M$, the rendered color of a pixel $x$ is the expected color along a camera ray $\bm{r}_{i,x}(t) = \bm{o}_i + t\bm{d}_{i,x}$, where $\bm{o}_i$ is the camera origin and $\bm{d}_{i,x}$ is the ray direction for pixel $x$ that can be computed from the corresponding camera pose $P_i$. The pixel value for 2D coordinate $x$ is then given by the line integral:
\begin{equation}
    \bm{C}_{\phi, \psi}\paren{\bm{r}_{i,x}} = \int_{t_{n}}^{t_f} \bm{\sigma}_{\psi}\paren{\bm{r}_{i,x}(t)}T\paren{\bm{r}_{i,x}(t)}\bm{c}_{\phi}\paren{\bm{r}_{i,x}(t)}\,dt,
\end{equation}
where $t_n, t_f$ defines the near and far plane, and 
\begin{equation}
T\paren{\bm{r}_{i,x}(t)} = \Exp\bracket{-\int_{t_n}^t\sigma\paren{\bm{r}_{i,x}(s)}\,ds}
\end{equation}
is the transmittance of the point $\bm{r}_{i,x}(t)$ along the direction $\bm{d}_{i,x}$.




\subsection{Implicit Maximum Likelihood Estimation}
\label{sec:imle}
One choice of probabilistic model is implicit maximum likelihood estimation (IMLE)~\cite{li2018implicit} that represents a distribution as a set of samples and is designed to handle possibly multimodal distributions.
As an implicit probabilistic model, IMLE learns a parameterized transformation $\bm{H}_{\theta}(\cdot)$ of a latent random variable, e.g. a Gaussian $\bm{z}\sim \mathcal{N}(0, \mathbf{I})$, where $\bm{H}_{\theta}(\cdot)$ often takes the form of a neural network that output samples 
$\bm{w}_j=\bm{H}_{\theta}(\bm{z}_j)$ with $\bm{w}_j \sim \mathbb{P}_{\theta}(\bm{w})$. Here, $\mathbb{P}_{\theta}$ is a probability measure obtained by transforming the standard Gaussian distribution measure via $\bm{H}_{\theta}$.
Given a set of data samples $\{\hat{\bm{w}}_1, ..., \hat{\bm{w}}_N\}$, the IMLE objective optimizes the model parameters $\theta$ with 





\begin{equation}
\label{eq:imle_obj}
    \hat{\theta} = \arg\min_{\theta}\E_{\bm{z}_1, \dots, \bm{z}_K}\bracket{\sum_{i=1}^N\min_{j}\norm{\bm{H}_{\theta}(z_j) - \hat{\bm{w}}_i}_2^2}.
\end{equation}
It is shown that the above objective to be equivalent to maximizing the likelihood. 


%% file: sections/4_method.tex
\section{Method}
\label{sec:method}
In the following sections, we formally define the provenance at all points as a stochastic field (Sec.~\ref{sec:method1}), extend IMLE to model the provenance field (Sec.~\ref{sec:method2}), and derive an equivalent pointwise loss for gradient descent (Sec.~\ref{sec:fimle}). 

\paragraph{Notations.} We denote a stochastic field with a calligraphic font ($\mathcal{D}_\theta$) and samples from the stochastic field using the same letter but bolded ($\bm{D}_\theta$). Concretely $\bm{D}_\theta$ is a function sample, a function defined over all points $\bm{x}\in \mathbb{R}^3$, that is sampled from the stochastic field $\mathcal{D}_\theta$, i.e. $\bm{D}_\theta\sim\mathcal{D}_\theta$. $\bm{D}_\theta$ maps each point to one possible sample in its provenances.
We also denote the distribution of the provenances at point $\bm{x}$ as $\mathcal{D}_\theta(\bm{x})$.
Moreover, a provenance sample $\bm{D}_\theta(\bm{x})$ from $\mathcal{D}_\theta(\bm{x})$ is equivalent to evaluating the function $\bm{D}_\theta\sim \mathcal{D}_\theta$ at $\bm{x}$. Finally, we let a hat ($\hat{\cdot}$) denote the empirical samples/distributions.
\subsection{Provenance as a Stochastic Field}
\label{sec:method1}



The \textbf{provenance} of a point is defined as the \emph{locations where it is likely visible from}, and as a point can be visible from multiple locations, it can be represented as samples from a distribution. Specifically, provenances of a point $\bm{x}$ can be represented as \emph{samples} from its provenance \emph{distribution} $\mathcal{D}_\theta(x)$. That means that the likelihood of sampling a location $\bm{y}\in \mathbb{R}^3$ from $\mathcal{D}_\theta(x)$ determines how likely $\bm{x}$ is visible from location $\bm{y}$. Because such distributions are defined for each 3D point $\bm{x} \in \mathbb{R}^3$, the collection of per-point provenances forms a stochastic field $\mathcal{D}_\theta$ indexed by coordinates $\bm{x}\in \mathbb{R}^3$.

Empirically, given sparse training camera views $P_1, \dots, P_M$, if $\bm{x}$ is inside the camera frustum $\Pi_i$ 
for view $P_i$ and is not occluded, an \emph{empirical} sample from the provenances of $\bm{x}$ can be parameterized as a distance-direction tuple $\hat{D}_i(\bm{x}) = (\hat{t}_{i, \bm{x}}, \hat{\bm{d}}_{i, \bm{x}}) \in \R_+ \times \mathbb{D}^3$~\footnote{$\mathbb{D}^3$ denotes a solid ball in $\mathbb{R}^3$}. Considering all $M$ training views, the empirical distribution of provenances at point $\bm{x}$ is defined by the following density function:
\begin{align}
\label{eq:empirical}
p_{emp}\paren{t, \bm{d}} = \frac{1}{M} \sum_{i=1}^M \delta[\paren{t, \bm{d}} &= \ \paren{t_{i,\bm{x}}, \bm{d}_{i, \bm{x}}}]\\
\label{eq:empirical2}
\text{where } \paren{t_{i,\bm{x}}, \bm{d}_{i, \bm{x}}} &= \paren{v_{i,\bm{x}}\norm{\bm{x}-\bm{o}_i}, v_{i,\bm{x}}\frac{\bm{x}-\bm{o}_i}{\norm{\bm{x}-\bm{o}_i}}}
\end{align}
where $\delta$ is the Dirac delta function; $v_{i, \bm{x}} \in [0, 1]$ determines the length of $\bm{d}$ to handle occlusions. We modeled it as the transmittance from the NeRF model. To recover the location that observes $\bm{x}$, we can write $\bm{y}_{i, \bm{x}} = \bm{x} - t_{i, \bm{x}}\bm{d}_{i, \bm{x}}$. 



While the above empirical distribution of provenances is given by the training cameras, the actual distribution of provenances, i.e. for each point the locations that point is likely visible from, can have a more complex dependence on both the underlying geometry and the cameras.
To capture this complexity, we model $\mathcal{D}_\theta\paren{\bm{x}}$ as a learnable network, a probabilistic model that can model potentially complex distribution, and one choice of such a model is implicit maximum likelihood estimate (IMLE)~\cite{li2018implicit}. Similar to the empirical distribution, we also represent provenance samples from $\mathcal{D}_\theta(\bm{x})$ as a distance-direction tuple $(t, \bm{d})$ as defined in Eq~\ref{eq:empirical2}. We optimized our network with the empirical distribution $\hat{\mathcal{D}}$ as training signals.


$\smash{\mathcal{D}_\theta(\bm{x}})$ defines a distribution for all point $\bm{x}\in \R^3$. Treating $\R^3$ as the index set, $\smash{\mathcal{D}_\theta} = \smash{\set{\mathcal{D}_\theta\paren{\bm{x}}}_{\bm{x}\in \R^3}}$  defines a stochastic field on $\R^3$ as a collection of distributions $\mathcal{D}_\theta\paren{\bm{x}}$ for all $\bm{x}\in \R^3$. Because a stochastic field is composed of infinitely many random variables over $\R^3$, existing methods cannot be applied out of the box as they only model finite-dimensional distributions. In the following sections, we extend IMLE~\cite{li2018implicit} to model this stochastic field.

\subsection{\methodname}
\label{sec:method2}
\begin{figure*}[t]
    \centering
    \includegraphics[width=\textwidth]{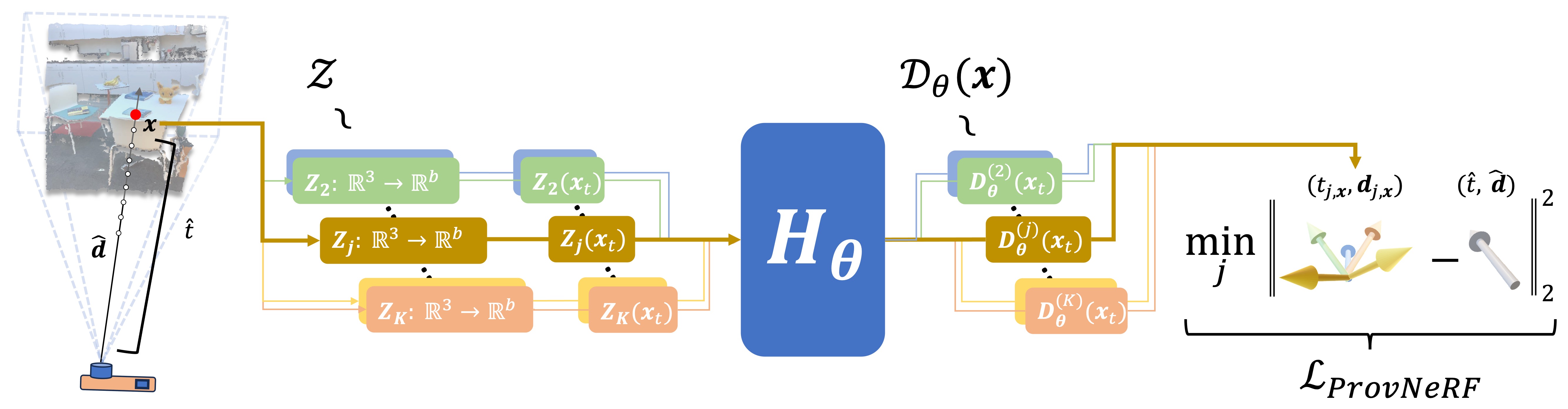}
    \caption{\textbf{Training pipeline for \methodname.} For each point $\bm{x}$ seen from provenance tuple $(\hat{t}, \hat{\bm{d}})$, with direction $\bm{d}$ at distance $t$, we first sample $K$ latent random functions $\{\bm{Z}_j\}$ from distribution $\mathcal{Z}$. The learned transformation $\bm{H}_\theta$ transforms each $\bm{Z}_j(\bm{x})$ to a provenance sample $\bm{D}_\theta^{(j)}(\bm{x})$. Finally $\bm{H}_{\theta}$ is trained with $\mathcal{L}_{\methodname}$ as defined in Eq.~\ref{eq:our_obj}.}
    \label{fig:pipeline}
\end{figure*}
\methodname models provenances of a NeRF as a stochastic field by extending IMLE~\cite{li2018implicit} to functional space. IMLE learns a mapping that transforms a latent distribution to the data distribution, where each data sample is either a scalar or a vector (Sec.~\ref{sec:imle}). However, in our context, since samples from the stochastic field $\mathcal{D}_\theta$ are \emph{functions} mapping 3D locations to provenances, we need to extend IMLE to learn a neural network mapping $\bm{H}_\theta$ that transforms a pre-defined \emph{latent stochastic field} $\mathcal{Z}$ to the provenance distribution $\mathcal{D}_\theta$ (See Fig.~\ref{fig:pipeline}). 

Let $\mathcal{Z}$ be the stochastic field where each sample $\bm{Z} \sim \mathcal{Z}$ is a function $\bm{Z}: \R^3 \to \R^b$. To transform $\mathcal{Z}$ to $\mathcal{D}_\theta$, fIMLE learns a deterministic mapping $\bm{H}_{\theta}$ that maps each latent function $\bm{Z} \sim \mathcal{Z}$ to a function $\bm{D}_\theta \sim \mathcal{D}_\theta$ via composition: $\bm{D}_\theta = \bm{H}_\theta\circ\bm{Z}$. $\bm{H}_\theta$ here is represented as a neural network to handle complex transformations from $\mathcal{Z}$ to $\mathcal{D}_\theta$.
We define a latent function sample $\bm{Z} \sim \mathcal{Z}$ to be the concatenation of a \emph{random linear transformation} of $\bm{x}$ and $\bm{x}$ itself. Mathematically, each latent function $\bm{Z} \sim \mathcal{Z}$ is a block matrix of size $(b + 4) \times 3$:
\begin{equation}
    \bm{Z}\paren{\bm{x}} = \bracket{\begin{array}{c}
\bm{z}  \\ \hline
\bm{I} 
\end{array}}\bm{x}, \text{ where }\;\bm{z} \sim \mathcal{N}\paren{\bm{0}, \lambda^2 \bm{I}}, \bm{x}\in \R^3.
\end{equation}
Although $\bm{Z}$ can be designed to have non-linear dependence on the input location $\bm{x}$, we experimentally show that this simple design choice works well across different downstream applications. 


To train $\bm{H}_\theta$, we maximize the likelihood of the training provenances (Eq.~\ref{eq:empirical}) under $\mathcal{D}_\theta$ for each $\bm{x}$ using the IMLE objective~\cite{li2018implicit} extended to functional space. We term this extension as functional Implicit Maximum Likelihood Estimation (fIMLE). Because a direct extension to fIMLE leads to an intractable objective, we derive an efficient pointwise loss between the training provenances and model predictions equivalent to the fIMLE objective in the following section.

\subsection{Functional Implicit Maximum Likelihood Estimation}
\label{sec:fimle}
\begin{figure}[t!]
\centering
\includegraphics[width=\linewidth]{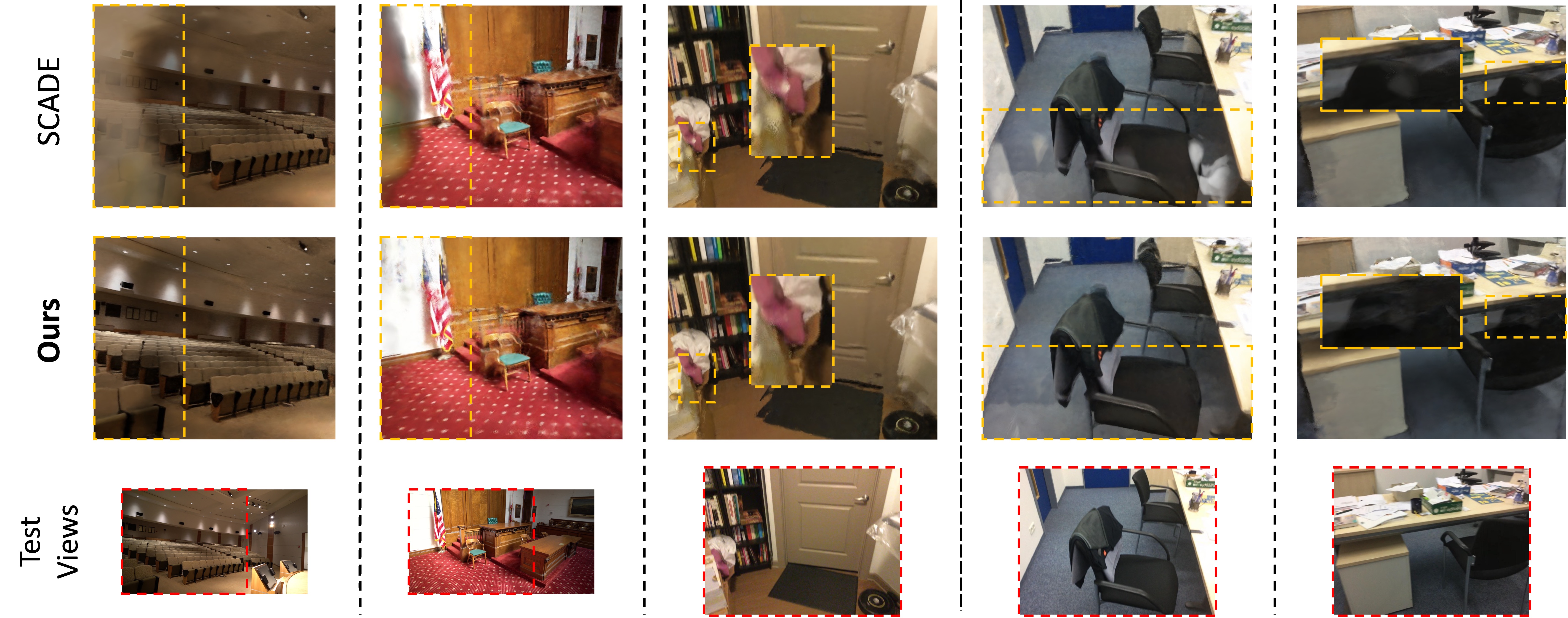}
\caption{ \textbf{Visual Effect of $\mathcal{L}_{\text{ProvNVS}}$ in Eq.~\ref{eq:nvs}}. Compared to pre-trained SCADE model, our method can remove additional floaters in the scene (see the boxed region). 
}
\label{fig:reg}
\end{figure}
\input{tables/reg_experiment}


\label{sec_experiments}

We construct an IMLE objective for stochastic fields to maximize the likelihood of training provenances under $\mathcal{D}_{\theta}$. Similar to Eq.~\ref{eq:imle_obj}, if we have i.i.d. empirical samples $\hat{\bm{D}}_1, \dots, \hat{\bm{D}}_M$ from the empirical stochastic field $\hat{\mathcal{D}}$ (defined in Sec.~\ref{sec:method1}), and model samples $\bm{D}_\theta^{(1)}, \dots, \bm{D}_\theta^{(K)}$ from the parameter stochastic field $\mathcal{D}_{\theta}$, we define the fIMLE objective as


\vspace{-0.3cm}
\begin{equation}
    \label{eq:obj}
    \begin{split}
        \hat{\theta} &= \arg\min_{\theta}\E_{\bm{D}_\theta^{(1)}, \dots, \bm{D}_\theta^{(K)}}
        \bracket{\sum_{i=1}^n\min_j\norm{\hat{\bm{D}}_i - \bm{D}_\theta^{(j)}}^2_{L^2}}.
    \end{split}
\end{equation}

Unlike the original IMLE objective (Eq.~\ref{eq:imle_obj}) that can be directly optimized, the fIMLE objective in Eq.~\ref{eq:obj} requires the computation of a $L^2$ integral norm -- a functional analogy to the $L^2$ vector norm -- which, in general, is not analytically in closed form. Furthermore, approximations of this integral are very expensive since each point query to $\bm{D}_\theta$ needs a forward pass through $\bm{H}_\theta$.



To get around this, we use the calculus of variations to reformulate Eq.~\ref{eq:obj} to minimize the pointwise difference between the empirical samples and model predictions~\footnote{See the supplementary for the full derivation}. This allows us to write the fIMLE objective as

\begin{equation}
    \label{eq:obj2}
    \mathcal{L}_{\text{fIMLE}} = \E_{\bm{D}^{(1)}_{\theta}, \dots, \bm{D}^{(K)}_{\theta} \sim \mathcal{D}_\theta} \bracket{\sum_{i=1}^n\min_j \E_{\bm{x}\sim \mathcal{U}\paren{\Omega}} \norm{\hat{\bm{D}}_i\paren{\bm{x}} - \bm{D}^{(j)}_{\theta}\paren{\bm{x}}}^2_2},
\end{equation}
where $\mathcal{U}(\Omega)$ is a uniform distribution over the scene bound $\Omega$. 
Eq.~\ref{eq:obj2} only requires computing the pointwise difference between samples from $\smash{\hat{\mathcal{D}}(\bm{x})}$ and $\smash{\mathcal{D}_{\theta}(\bm{x})}$, making it efficiently optimizable with gradient descent. 
Ultimately, \methodname jointly updates the underlying NeRF's parameters and $\smash{\mathcal{D}_\theta}$ by minimizing
\begin{equation}
    \label{eq:our_obj}
        \mathcal{L}_{\text{\methodname}} = \mathcal{L}_{\text{NeRF}} + \mathcal{L}_{\text{fIMLE}} = \mathcal{L}_{\text{NeRF}} + \E_{\bm{D}^{(1)}_{\theta}, \dots, \bm{D}^{(K)}_{\theta}, \bm{x}} \bracket{\min_j\E_{\paren{\hat{t},\hat{\bm{d}}}}\norm{(\hat{t}, \hat{\bm{d}}) - \paren{t_{j, \bm{x}}, \bm{d}_{j, \bm{x}}}}^2_2}
\end{equation}
where $\paren{t_{j, \bm{x}}, \bm{d}_{j, \bm{x}}} = \bm{D}^{(j)}_\theta\paren{\bm{x}}$, $(\hat{t},\hat{\bm{d}})$ are i.i.d. samples from $\hat{\mathcal{D}}(\bm{x})$, and $\mathcal{L}_{\text{NeRF}}$ 
is the original objective of the NeRF model, e.g. photometric loss and depth loss.
We provide implementation and architectural details in the supplementary material. See Figure~\ref{fig:pipeline} for the training pipeline illustration. 

%% file: tables/reg_experiment.tex
\begin{table}[th]
    \centering
    \begin{tabular}{lccc|ccc}
    \toprule
        & \multicolumn{3}{c}{\textbf{Scannet}} & \multicolumn{3}{c}{\textbf{Tanks and Temple}} \\ 
         & PSNR $\paren{\uparrow}$ & SSIM $\paren{\uparrow}$ & LPIPS $\paren{\downarrow}$ & PSNR $\paren{\uparrow}$ & SSIM $\paren{\uparrow}$ & LPIPS $\paren{\downarrow}$  \\
         \midrule
         NeRF~\cite{mildenhall2020nerf} & 19.03 & 0.670 & 0.398 & 17.19 & 0.559 & 0.457\\
        DDP~\cite{roessle2022depthpriorsnerf} & 19.29 & 0.695 & 0.368 & 19.18 & 0.651 & 0.361\\
         SCADE~\cite{scade}& \underline{21.54} & 0.732 &\underline{0.292} & \underline{20.13} & \underline{0.662} & \underline{0.358} \\
         DäRF~\cite{song2023d}& 21.28 & \textbf{0.741} &0.323 & 19.67 &0.652 &0.374\\
         \midrule
        \textbf{Ours} & \textbf{21.73} & \underline{0.733} &\textbf{0.291} & \textbf{20.36} & \textbf{0.663} & \textbf{0.349}\\
        
        \bottomrule
    \end{tabular}
    \caption{\textbf{Novel View Synthesis Results.} Our method outperforms baselines in novel view synthesis on both Scannet and Tanks and Temple Datasets. This is because our novel NeRF regularizer in Eq.~\ref{eq:nvs} can remove additional floaters in the scene as shown in Fig.~\ref{fig:reg}. See Sec.~\ref{sec_nvs} for details.}
    \label{tab:reg_exp}
\end{table}

%% file: sections/5_results.tex
\section{Experiments}
Our ProvNeRF learns per-point provenance field $\mathcal{D}_\theta$ by optimizing $\mathcal{L}_{\text{ProvNeRF}}$ on a NeRF-based model. To validate ProvNeRF, we demonstrate that jointly optimizing the provenance distribution and NeRF representation can result in better scene reconstruction as shown in the task of novel view synthesis (Sec.~\ref{sec_nvs}). Moreover, we also show that the learned provenance distribution enables other downstream tasks such as estimating the uncertainty of the capturing field (Sec.~\ref{sec_uncertainty}). We provide an ablation study on fIMLE against other probabilistic methods in Sec.~\ref{sec_ablation}. 




\noindent \textbf{Stochastic Provenance Field Visualization} Fig.~\ref{fig:direction_vis} visualizes the provenance stochastic field by sampling $16$ provenances on a test view of the Scannet 758 scene. The directions of the samples are the \textit{negative} of the predicted provenance directions for better illustration. Each sample is colored based on its predicted visibility. Notice that fIMLE allows ProvNeRF to predict multimodal provenance distributions at different scene locations.  
\subsection{Novel View Synthesis}
\label{sec_nvs}

We show modeling per-point provenance improves sparse, unconstrained novel view synthesis. As a point's provenances are sample locations from where the point is likely visible, the region between the provenance location samples and the query point should likely be empty. We design our provenance loss for NVS with this intuition.

\input{tables/ablation_nvs}
Concretely, starting from a given NeRF model,  we first sample points $\bm{x}_1, ..., \bm{x}_N$ for a training camera ray parameterized as $\hat{\bm{r}}_{x}(t)$. Here we denote point $\bm{x}_i=\hat{\bm{r}}_{x}(\hat{t}_i)$. We only take points $\bm{x}_i$ with transmittance greater than a selected threshold $\lambda=0.9$. For each visible point $\bm{x}_i$, we sample provenances $\smash{(t_1^{(i)}, \bm{d}_1^{(i)})}, \dots, \smash{(t_K^{(i)}, \bm{d}_K^{(i)})}$ from $\mathcal{D}_\theta(\bm{x}_i)$ with $\lVert{\bm{d}_1^{(i)}\lVert}_2 \geq 0.7$. Then each distance-direction tuple $\smash{(t_j^{(i)}, \bm{d}_j^{(i)})}$ gives a location $\smash{\bm{y}_j^{(i)} = \bm{x}_i - t_j^{(i)}\bm{d}_j^{(i)}}$ from which $\bm{x}_i$ is observed. This in turn means $\bm{x}$ should be equally visible when rendered from ray parameterized as $\smash{\bm{r}_{x}^{(i)}(t) = \bm{y}_j^{(i)} + t\bm{d}_j^{(i)}}, \forall j$. 
With this, we define our provenance loss for novel view synthesis as
\begin{equation}
\label{eq:nvs}
    \mathcal{L}_{\text{ProvNVS}} = \sum_{i=1}^N \sum_{j=1}^K \bracket{ \alpha\; +\; T(\bm{r}_x^{(i)}(t_j^{(i)})) - T(\hat{\bm{r}}_x(\hat{t}_i))}_+,
\end{equation}
\noindent where $[\dots]_+$ denotes the hinge loss and $\alpha=0.05$ is a constant margin. $\mathcal{L}_{\text{ProvNVS}}$ encourages the transmittance at $\bm{x}_i$ along training camera rays to be \emph{at least} the visibility predicted by the sampled provenances from the provenance field with margin $\alpha$. By matching transmittances between the provenance directions and the training rays, $\mathcal{L}_{\text{ProvNVS}}$ can be used together with $\mathcal{L}_{\methodname}$ to optimize the NeRF representation and the provenance field, resulting in an improved scene geometry. We apply ProvNeRF to SCADE~\cite{scade} for the task of novel view synthesis. See the supplement for details on the dataset, metrics, baselines, and implementation details.


\noindent \textbf{Results.}
Table~\ref{tab:reg_exp} shows our approach outperforms the state-of-the-art baselines in NVS on scenes from both the Scannet~\cite{dai2017scannet} and Tanks and Temples~\cite{Knapitsch2017} dataset. Qualitative comparisons are shown in Fig.~\ref{fig:reg}. We see that compared to the baseline SCADE, whose geometry is already relatively crisp, our $\mathcal{L}_{\text{ProvNVS}}$ can further improve its NVS quality by removing additional cloud artifacts, as shown in the encircled regions. Note that this improvement does not require any additional priors and is only based on the provenance of the scene. 

We also compare our performance with deterministic baselines: \textit{Deterministic Field} regresses one provenance for each 3D location using a neural network and \textit{Frustum Check} calculates the training provenance defined in Eq.~\ref{eq:empirical} by back-projecting the sampled points to one of the training camera and use that as the regularization information. Table~\ref{tab:ablation_nvs} shows that our provenance field outperforms these baselines on the novel view synthesis task because the deterministic field cannot model complex provenance distribution and the frustum check baseline lacks generalization ability as it cannot be optimized to adapt the output provenance based on the current NeRF's geometry. 



\begin{figure*}[t!]
    \centering
    \includegraphics[width=\textwidth]{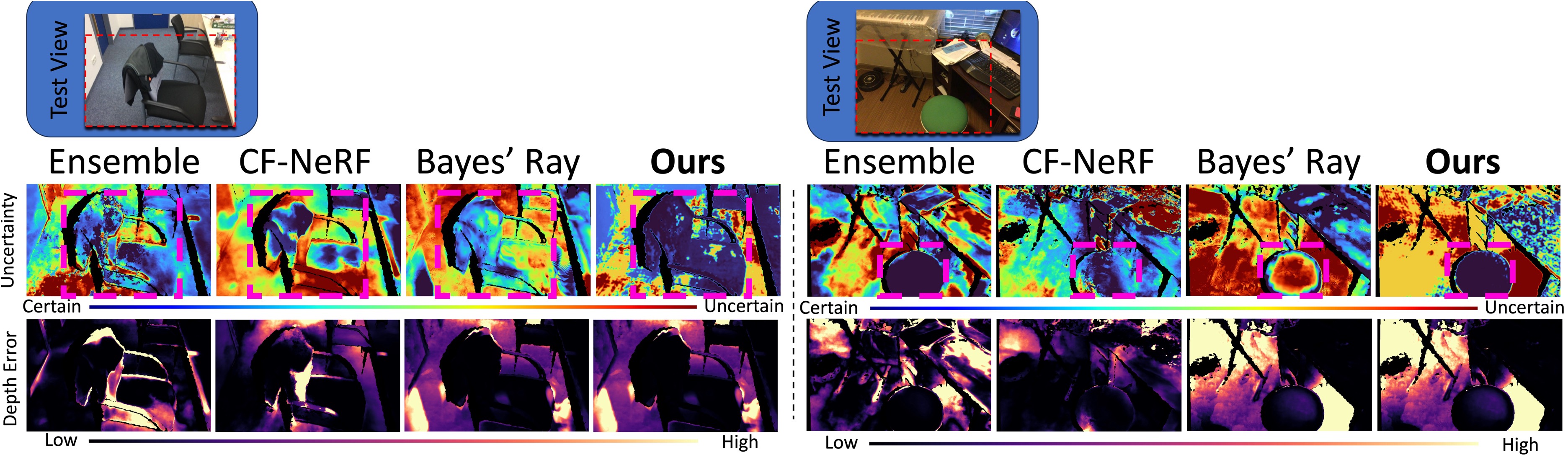}
    \caption{\textbf{Qualitative Results for Uncertainty Modeling.} We visualize our uncertainty maps obtained using the method described in Sec.~\ref{sec_uncertainty}. The uncertainty and depth error maps are shown with color bars specified. 
    Uncertainty values and depth errors are normalized per test image for the result to be comparable. As shown in the boxed regions, our method predicts uncertainty regions with more correlation with the predicted depth errors. }
    \label{fig:unc}
\end{figure*}
\input{tables/unc_nll_experiment}
\vspace{-0.3cm}
\subsection{Modeling Uncertainty in the Capturing Process}
\label{sec_uncertainty}
\begin{wrapfigure}{r}{0.47\textwidth}
  \begin{center}
    \includegraphics[width=\textwidth]{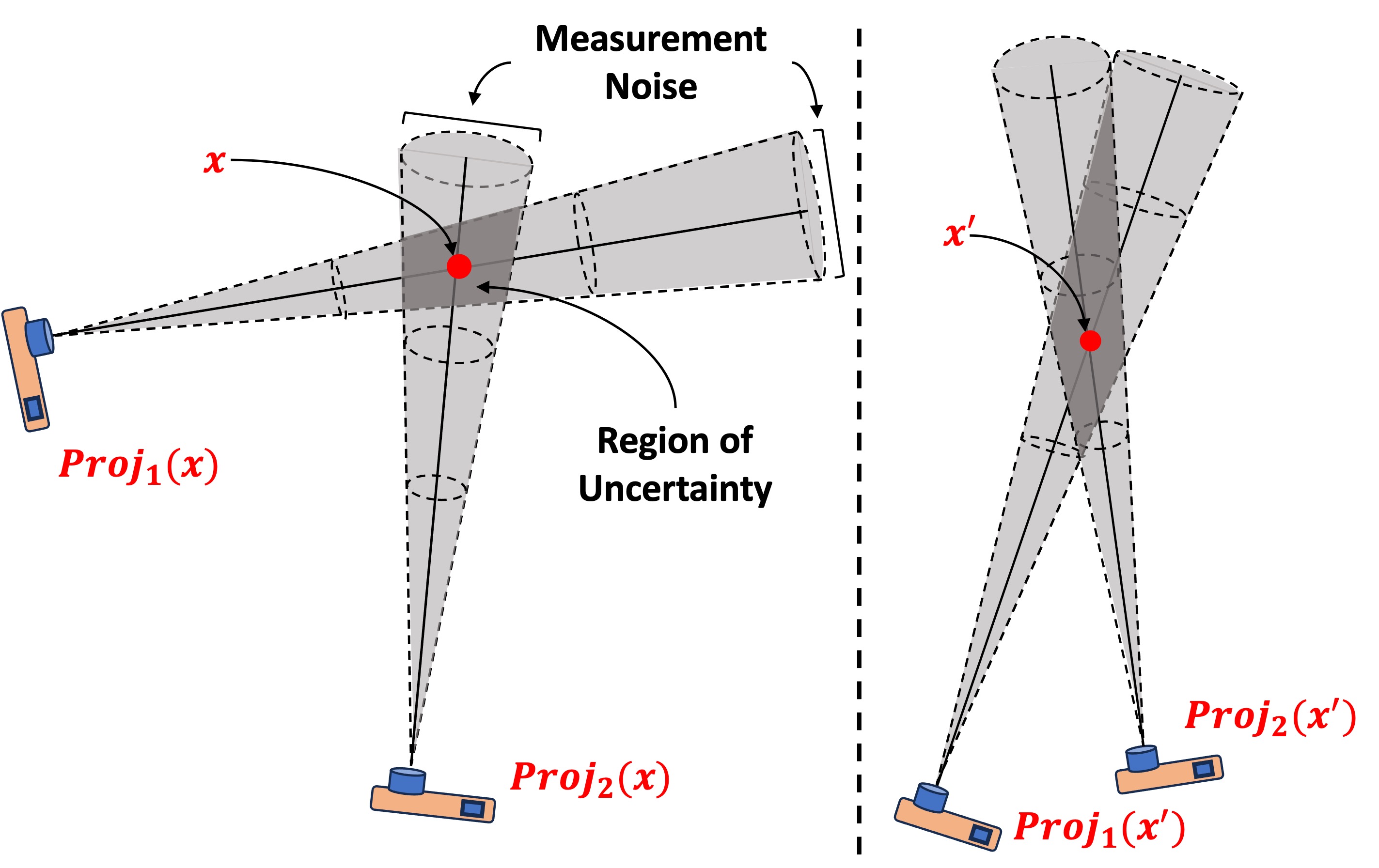}
  \end{center}
  \caption{\small{\textbf{Triangulation Uncertainty~\cite{hartley_zisserman_2004}.} The figure shows that $\bm{x}^\prime$ is more uncertain compared to $\bm{x}$ because the predicted provenances for $\bm{x}^\prime$ give a narrower baseline than the baseline given by provenances of $\bm{x}$.}}
    \label{fig:tri}
\end{wrapfigure}
Provenances allow for estimating the uncertainty in triangulation, i.e., the capturing process. In classical multiview geometry~\cite{hartley_zisserman_2004}, the angle between the rays is a good rule of thumb that determines the accuracy of reconstruction. Fig.~\ref{fig:tri} illustrates this rule as the region of uncertainty changes depending on the setup of the cameras. Formally, for a 3D point $\bm{x}$, we sample provenances $\set{(t_j, \bm{d}_j)}^K_{j=1}$ from $\mathcal{D}_\theta(\bm{x})$. Treating $\bm{d}_j$ as the principal axes and $t_j$ as the distances from $\bm{x}$ to the camera origin, each provenance sample defines a pseudo camera $P_j$ that observes $\bm{x}$ at pixel location $x_j = \text{Proj}_j(\bm{x})$. Following chapter 12.6 of~\cite{hartley_zisserman_2004}, we define the triangulation uncertainty of $\bm{x}$ as the probability of $\bm{x}$ given its noisy 2D pseudo observations: $\mathbb{P}\paren{\bm{x} | x_1, \dots, x_K} \propto \mathbb{P}\paren{x_1, \dots, x_K |  \bm{x}}\mathbb{P}\paren{\bm{x}} = \prod_{j=1}^K\mathbb{P}\paren{x_j |  \bm{x}}\mathbb{P}\paren{\bm{x}}.$

The last two equalities are derived by assuming independence of the 2D observations and each $\mathbb{P}(x_j| \bm{x})$ follows a Gaussian distribution $\smash{\mathcal{N}(\text{Proj}_j(\bm{x}), \sigma^2)}$. This assumption is equivalent to corrupting each 2D observation $\text{Proj}_j(\bm{x})$ by a zero-mean Gaussian noise with $\sigma^2$ variance, accounting for measurement noises in the capturing process. Assuming a uniform prior of $\mathbb{P}(\bm{x})$ over the scene bound, the exact likelihood can be efficiently computed with importance sampling. This quantifies a point's triangulation quality given the sampled provenances, which becomes a measurement of the uncertainty of the capturing process. We apply our provenance field to ProvNeRF with different NeRF backbones~\cite{scade, roessle2022depthpriorsnerf} and compute the likelihood. See supplementary for details on the dataset, metrics, baselines, and implementations.  
\vspace{-0.3cm}
\paragraph{Results.} Tab.~\ref{tab:unc_nll} shows the quantitative results on Scannet~\cite{dai2017scannet} and Matterport3D~\cite{Matterport3D}. We follow~\cite{s-nerf,sünderhauf2022densityaware} to measure the negative log-likelihood (NLL) of the ground-truth surface under each model's uncertainty prediction. Since our \methodname can be applied to any pre-trained NeRF module, we use pre-trained SCADE~\cite{scade} for Scannet and DDP~\cite{roessle2022depthpriorsnerf} for Matterport3D, both of which are state-of-the-art approaches in each dataset. Our approach achieves the best NLL across all scenes in both datasets by a margin because we compute a more fundamentally grounded uncertainty from classical multiview geometry~\cite{hartley_zisserman_2004} based on triangulation errors, while both CF-NeRF and Bayes' Rays require an approximation of the true posterior likelihood. Fig.~\ref{fig:unc} shows qualitative comparisons between baselines' and our method's uncertainty estimation. We expect a general correlation between uncertain regions with high-depth errors. An ideal uncertainty map should mark high-depth error regions with high uncertainty and vice versa. As shown in the boxed regions in the figure, our method's uncertainty map shows better correlation with the depth error maps. We also quantitatively evaluate uncertainty maps using negative log-likelihood following prior works Notice that in both examples, our method's certain (blue) regions mostly have low-depth errors (e.g., encircled parts in Fig.~\ref{fig:unc}) because our formulation only assigns a region to be certain if it is well triangulated (Fig.~\ref{fig:tri}). On the other hand, baselines struggle in these regions because they either use an empirical approximation from data or a Gaussian approximation of the ground truth posterior likelihood.

\begin{figure}
\begin{floatrow}
\ffigbox{%
  \includegraphics[width=0.45\textwidth]{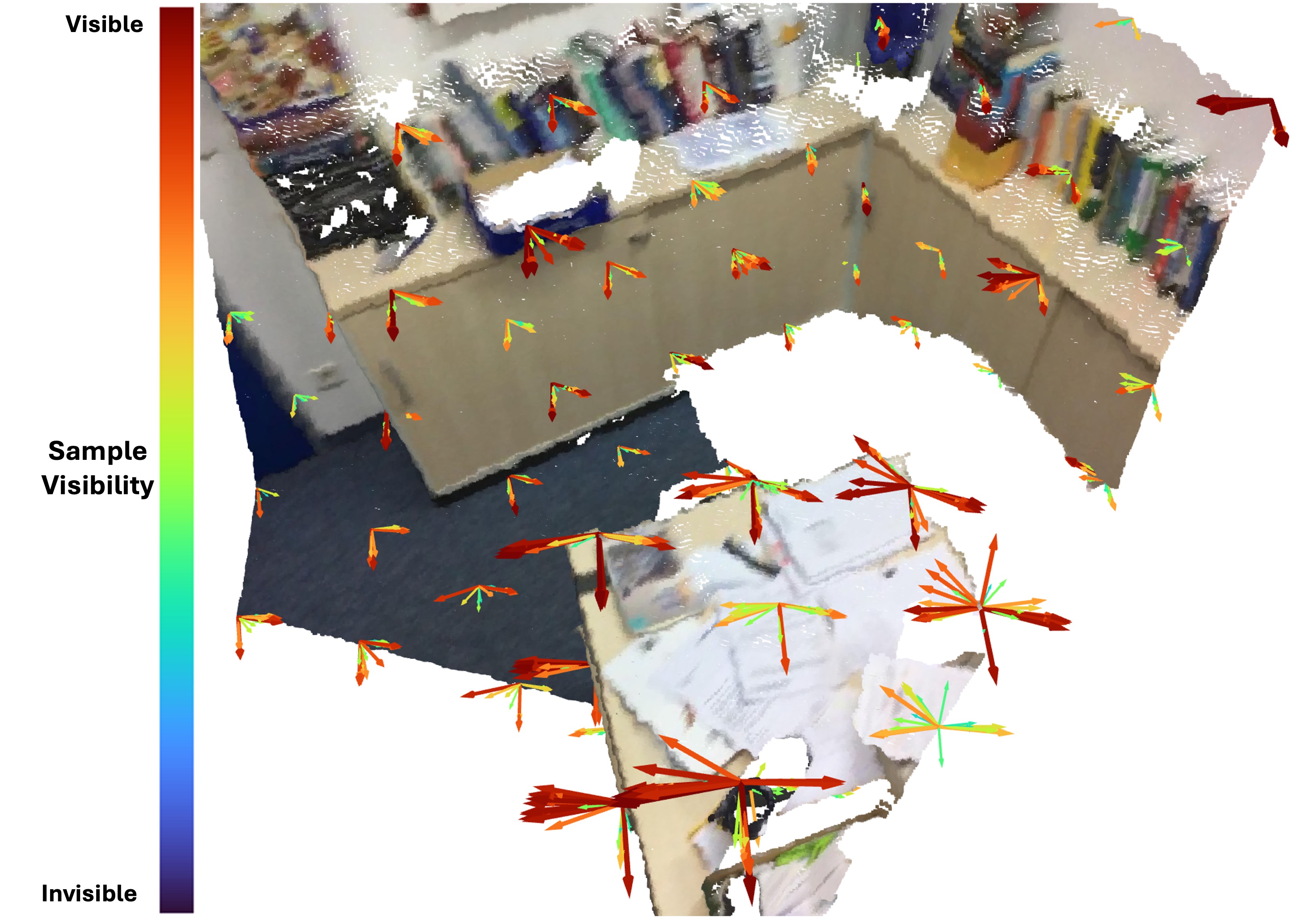}%
}{%
  \caption{\small{\textbf{Visualization of Provenance Field.}}}%
  \label{fig:direction_vis}
}
\capbtabbox{%
\resizebox{\linewidth}{!}{
  \begin{tabular}{lcc}
  \toprule
         & AP $\paren{\uparrow}$ & AUC $\paren{\uparrow}$ \\
         \midrule
         \small{Deterministic Field} & 0.163 &0.168 \\
        \small{Gaussian-based w/} $C = 2$ & 0.537 & 0.539\\
        \small{Gaussian-based w/} $C = 5$& 0.629 & 0.631\\
        \small{VAE-based} & 0.323 & 0.325\\
        \midrule
        \small{\methodname w/ Spatial Inv.} $\mathcal{Z}$& \underline{0.742}	& \underline{0.744}\\
        \textbf{Ours} & \textbf{0.745} & \textbf{0.747} \\
        \bottomrule
    \end{tabular}
    }
}{%
  \caption{\small{\textbf{Ablation Results on Scannet.}}}%
  \label{tab:ablation}
}
\end{floatrow}
\end{figure}
\vspace{-0.3cm}
\subsection{Preliminary Extension to 3D Gaussian Splatting} 
Because ProvNeRF is a post-hoc method that can model the provenance information for arbitrary novel view synthesis representations, we conduct a preliminary experiment that extends our provenance field modeling to 3D Gaussian Splatting~\cite{kerbl3Dgaussians}. Specifically, given a pre-trained Gaussian representation $\mathcal{G}$, we model a provenance distribution for each splat using IMLE with a shared 6-layer MLP for $\bm{H}_\theta$. The post-training takes around 30 minutes on a single A6000 Nvidia GPU. The optimized stochastic provenance field can then be queried at the Gaussian locations for the provenance samples. To show the usefulness of ProvNeRF applied to 3DGS, we use the methodology in Sec.~\ref{sec_uncertainty} to estimate uncertainty maps and compare them with the predicted depth errors. Fig.~\ref{fig:3dgs-unc} shows a qualitative comparison of our uncertainty map w.r.t. FishRF~\cite{jiang2023fisherrf}, a recent 3DGS uncertainty estimation baseline. Compared to their uncertainty map, ours shows more correlation to the depth error as highlighted by the boxed regions. Quantitatively we evaluate NLL on the three Scannet scenes shown on the right side of the same figure and show substantial improvements over FishRF. This improvement over existing literature suggests applying ProvNeRF to other representations such as 3DGS is promising. We leave further exploration of the method and applications as future works.
\subsection{Ablation Study}
\label{sec_ablation}
We validate the choice of fIMLE as our probabilistic model by measuring the average precision (AP)~\cite{voc} and area under the curve (AUC)~\cite{voc} of predicted provenances $(t, \bm{d})$ against ground truth provenances $(\hat{t}, \hat{\bm{d}})$ for a set of densely sampled points in the scene bound. See supplementary for metric and ablation implementation details.
\vspace{-0.3cm}



\paragraph{Deterministic v.s. Stochastic Field.} We validate the importance of modeling per-point provenance as a stochastic field rather than a deterministic field. We model $\mathcal{D}_\theta$ with a deterministic field parameterized by a neural network. Table~\ref{tab:ablation} shows the importance of modeling per-point provenance as a stochastic field. Since the provenances of a point are inherently multimodal, a deterministic field that only maps each $\bm{x}$ to a single provenance cannot capture this multimodality. 

\vspace{-0.3cm}

\paragraph{Choice of Probabilistic Model.} We validate our choice of fIMLE~\cite{li2018implicit} as our probabilistic model. We first compare with explicit probabilistic models that model the provenance field as a mixture of $C$ Gaussian processes and a VAE-based model. Table~\ref{tab:ablation} shows results for the Gaussian Mixture field with $C = 2, 5$ and the VAE-based process. Although the performances for the Gaussian-based models improve as we increase $C$, they still suffer from expressivity because of their explicit density assumption. Similarly, the VAE-based model suffers from mode-collapse while our fIMLE enables capturing a more complex distribution with a learned transformation $\bm{H}_{\theta}$.

\vspace{-0.3cm}

\paragraph{Choice of Random Function $\mathcal{Z}$.} Lastly, we validate our latent stochastic field $\mathcal{Z}$. We ablate our choice of $\mathcal{Z}$ with instead using a spatially \emph{invariant} latent stochastic field $\mathcal{Z}^\star$ with $\mathcal{Z}^\star\paren{\bm{x}} = \bracket{\bm{\varepsilon}, \bm{x}} \forall x$. Here, $\bm{\varepsilon}$ is a Gaussian noise vector in $\R^d$. Table~\ref{tab:ablation} shows the comparison between $\mathcal{D}_\theta$ obtained by transforming $\mathcal{Z}$ (\textbf{Ours}) and transforming $\mathcal{Z}^\star$ (\textbf{Spatial Inv.} $\mathcal{Z}$). We see that using a spatially varying latent stochastic field further increases the expressivity of our model.

%% file: tables/ablation_nvs.tex
\begin{wraptable}{r}{8cm}
    \begin{tabular}{lccc}
    \toprule
         & \small{PSNR} $\paren{\uparrow}$ & \small{SSIM} $\paren{\uparrow}$ & \small{LPIPS} $\paren{\downarrow}$\\
         \midrule
        \small{Determinisic Field}  & 21.38 & 0.720 & 0.307 \\
        \small{Frustum Check} & 21.56 & 0.728 & 0.297\\
        \midrule
        \textbf{Ours} & \textbf{21.73} & \textbf{0.733} & \textbf{0.291} \\
        \bottomrule
    \end{tabular}
    \caption{\textbf{NVS Ablation Results on Scannet.}}
    \label{tab:ablation_nvs}
\end{wraptable}


%% file: tables/unc_nll_experiment.tex
\begin{table*}[t]
    \centering
    \begin{tabular}{lcccccccc}
    \toprule
         & \multicolumn{4}{c}{\textbf{Scannet}} & \multicolumn{4}{c}{\textbf{Matterport}} \\ 
         \cmidrule(lr){2-5}\cmidrule(lr){6-9}
         & Avg. & $\#\text{710}$ & $\#\text{758}$ & $\#\text{781}$ & Avg. & Room 0 & Room 1& Room 2  \\
         \midrule
         Ensemble & 7.71 & \underline{3.01} & \underline{2.96} & 17.2 & 63.0 & 8.04 & 110 & 71.3\\
         CF-NeRF~\cite{CF-NeRF} & 660 & 430 & 571 & 980 & 507 & 799 & 488 & 233\\
         Bayes' Rays~\cite{bayesrays}& \underline{5.47}& 5.11 & 5.23 & \underline{6.07} & \underline{5.49} & \underline{5.67} & \underline{5.77} & \underline{5.91} \\
         \midrule
        \textbf{Ours} & \textbf{-3.05} & \textbf{0.19} & \textbf{-1.93} &\textbf{-7.40} & \textbf{-11.0} & \textbf{-13.6}&\textbf{-10.2} &\textbf{-9.17} \\
        \bottomrule

    \end{tabular}
    \caption{\textbf{NLL Results for Triangulation Uncertainty.}\vspace{-0.7cm} }
    \label{tab:unc_nll}
\end{table*}


%% file: sections/6_conclusion.tex
\begin{figure*}
    \centering
    \includegraphics[width=\textwidth]{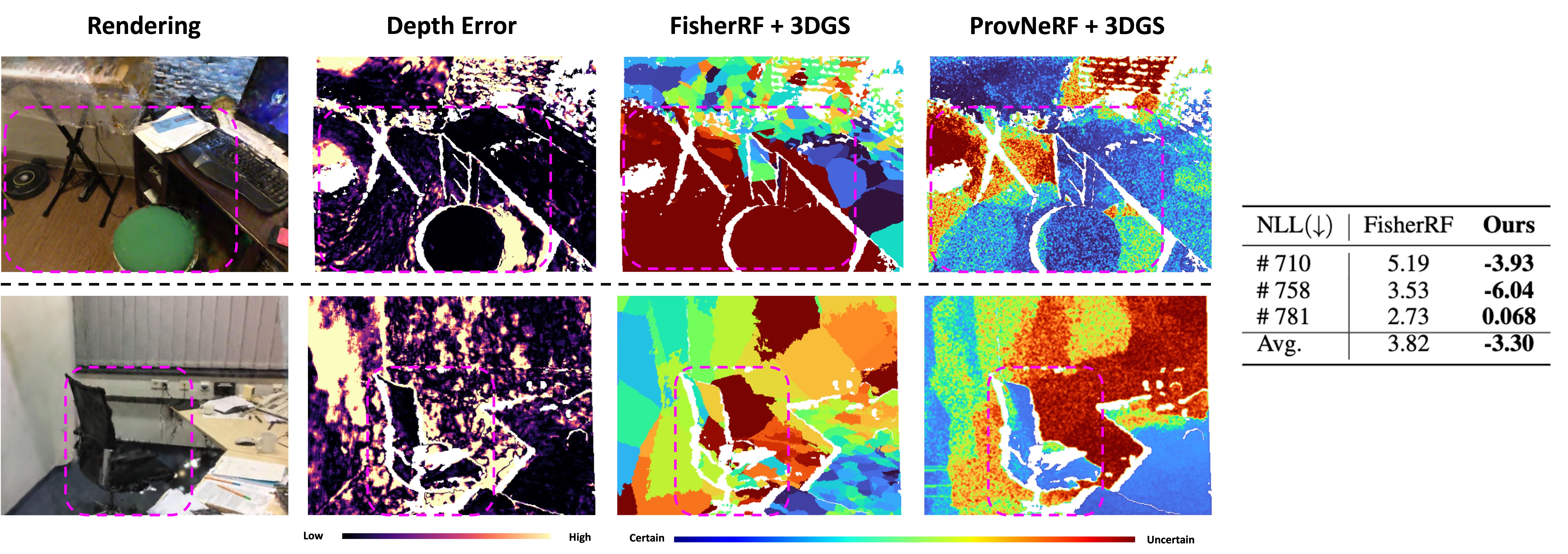}
    \caption{\textbf{Uncertainty Estimation Comparison with 3DGS.} Compared with FishRF, our method is able to estimate uncertainties that correlate more with the depth error as shown by the encircled regions. The right shows a quantitative comparison of uncertainty in negative log-likelihood. We outperform FisherRF in all three scenes.}
    \label{fig:3dgs-unc}
\end{figure*}
\section{Conclusion, Limitation, \& Future Works}
\label{sec:conclusion}
\vspace{-0.3cm}

We present ProvNeRF, a model that enhances the traditional NeRF representation by modeling provenance through an extension of IMLE for stochastic processes. ProvNeRF can be easily applied to any NeRF model to enrich its representation. We showcase the advantages of modeling per-point provenance in various downstream applications such as improving novel view synthesis and modeling the uncertainty of the capturing process. 

We note that our work is not without limitations. Our ProvNeRF requires post-hoc optimization which takes around 8 hours limiting its current usability for real-time or on-demand applications. The idea presented in our work is however not specific to the model design and it can be adapted to other representations. 

We also note that the hyperparameters to incorporate ProvNeRF are chosen for better performance, e.g. for the uncertainty and novel view synthesis applications, and in the future, it will be beneficial to explore a more adaptive approach in integrating provenance to different downstream applications. \\\\

%% file: sections/7_acknowledgement.tex
\noindent\textbf{Acknowledgement} This work is supported by a Vannevar Bush Faculty Fellowship, ARL grant W911NF-21-2-0104, an Apple Scholars in
AI/ML PhD Fellowship, and the Natural Sciences and Engineering Research Council of Canada (NSERC).

%% file: sections/X_suppl.tex
\clearpage
\setcounter{section}{0}
\renewcommand{\thesection}{S\arabic{section}}%
\setcounter{page}{1}
\numberwithin{equation}{section}
\setcounter{table}{0}
\renewcommand{\thetable}{S\arabic{table}}%
\setcounter{figure}{0}
\renewcommand{\thefigure}{S\arabic{figure}}%
\title{\methodname : Modeling per Point Provenance in NeRFs as a Stochastic Field -- Appendix}
\author{}
\date{}

\maketitle
We provide additional results and visualizations and elaborate on implementation, metrics, baselines, and dataset details for \methodname in this supplementary material. We organize the supplement as the following: in Sec.~\ref{sec:my_detail}, we detail the implementation detail of \methodname. In Sec.~\ref{sec:supp_unc}, we elaborate on the metrics, ablation, and implementation details and further provide additional visualizations on both the Scannet and Matterport datasets for our uncertainty modeling application. In Sec.~\ref{viewpoint}, we provide an additional application using our provenance field that selects viewpoints given different criteria. In Sec.~\ref{sec:nvs_imp}, we provide implementation details on novel view synthesis and further ablation results against other probabilistic methods. In Sec.~\ref{sec:ablation_supp}, we conduct further ablation studies and explain the implementation details of the ablations introduced in the main paper. We also further provide more details on the derivation of our fIMLE objective in Sec~\ref{sec_supp_fimle}. Finally, in Sec~\ref{sec:impact}, we discuss the potential societal impact of our work on climate change. 
\section{Provenance, Uncertainty, and NVS Visualization}
We attach a video showing 3D visualization of provenances, triangulation uncertainty, and novel view synthesis results. 
\section{\methodname Implementation Details}
\label{sec:my_detail}
Since our method can be plugged into any NeRF model, we apply our method to SCADE~\cite{scade} and DDP~\cite{roessle2022depthpriorsnerf}, two recent state-of-the-art sparse, unconstrained NeRF models. On each of the backbones, we append an additional provenance prediction branch that is a 3-layer MLP. The provenance prediction branch takes the output of the latent random functions $\bm{Z}\paren{\bm{x}}$ and outputs a provenance tuple $\paren{t_{\bm{x}}, \bm{d}_{\bm{x}}}$ as defined in the main. In practice, the provenance branch outputs $t_{\bm{x}}$ as a normalized distance by the near and far plane of the training cameras, and the direction $\bm{d}_{\bm{x}}$ is a 3D vector with its norm indicating the visibility/transmittance of $\bm{x}$ when seen from $\bm{y} = \bm{x} - t\bm{d}_{\bm{x}}$. 

Fig.~\ref{fig:pipeline} visualizes an iteration of the provenance distribution update. To train the distribution $\mathcal{D}_\theta$, we initialize \methodname the weights of the pre-trained NeRF model and run stochastic gradient descent on the provenance prediction branch using our $\mathcal{L}_{\text{ProvNeRF}}$ loss (defined in Eq. 11 of the main paper) for $200$K iterations. During the first 100K iterations we freeze the NeRF model and only update the provenance prediction branch. In each training iteration, we pick a training camera pose at random and shoot $256$ rays within the image uniformly. The loss in Eq. 10 in main is then computed and backpropagated for all coarse and fine samples along the ray against the direction and camera distance from the rays. The set of random functions is resampled every 1000 iterations. The size of the provenance function pool is set to $K=16$ and the latent space for latent random functions $\mathbb{Z}$ is set to $b = 32$. Additionally, following the usual NeRF training, the input and output of the latent random function both use the positionally encoded $\bm{x}$ instead of the actual 3D coordinate. The entire training process takes around $8$ hours on an NVIDIA A5000. Once $\mathcal{D}_\theta$ is trained, provenances of each point $\bm{x}$ can be sampled by sampling random functions $\bm{Z}$ from the distribution $\mathcal{Z}$ and evaluate $\bm{H}_\theta\circ\bm{Z}\paren{\bm{x}}$. To obtain the visible provenances, we only use samples with the norm of the predicted directions $\bm{d}_{\bm{x}} > 0.7$.  

\section{Uncertainty Estimation}
\label{sec:supp_unc}

\begin{figure*}[hbtp]
\centering
\includegraphics[width=0.9\textwidth]{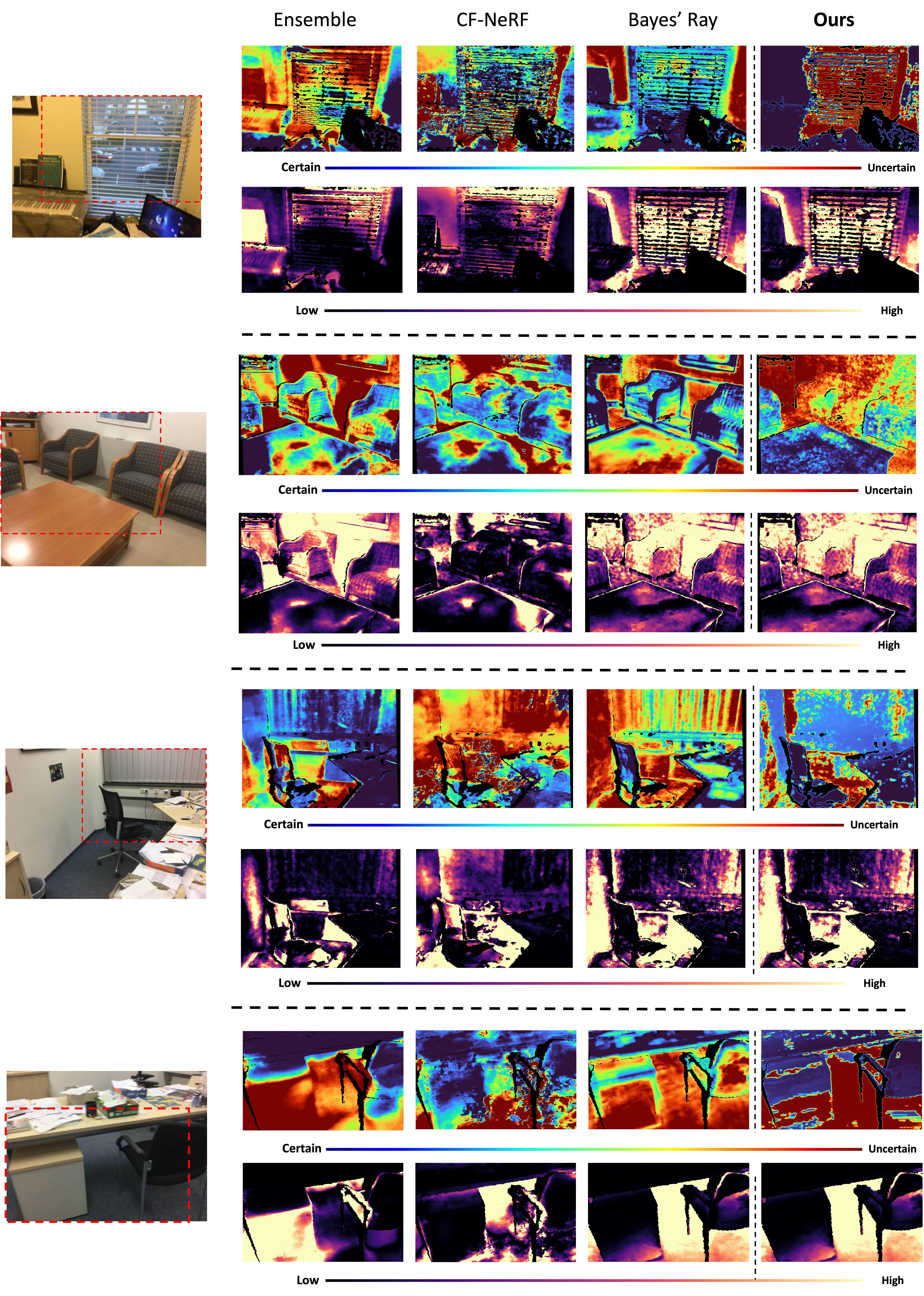}
\caption{ \textbf{Visualization of Uncertainty Estimation on the Scannet Dataset}. 
}
 \vspace{5em}
\label{fig:scannet_supp}
\end{figure*}
\begin{figure*}[hbtp]
\centering
\includegraphics[width=\textwidth]{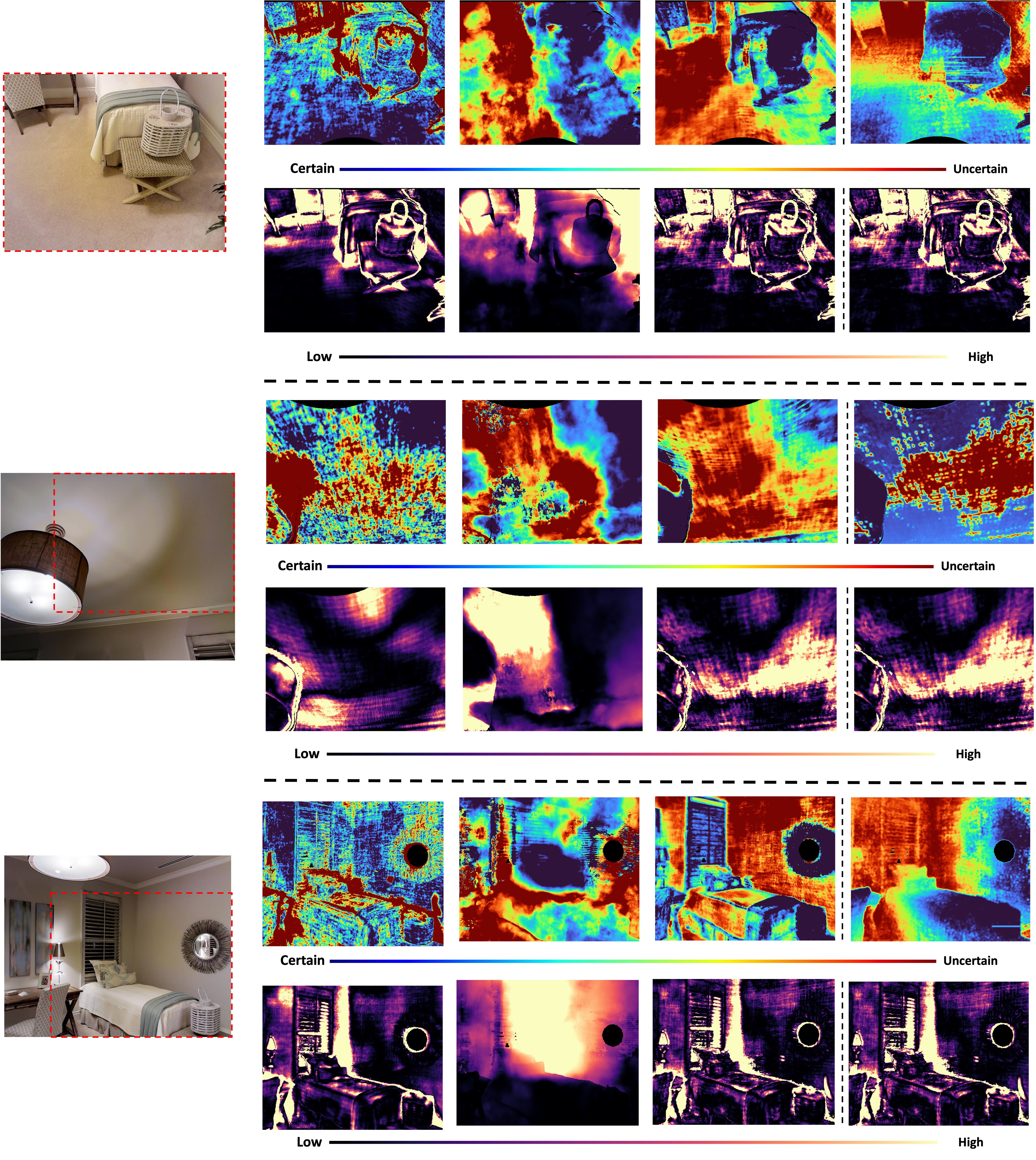}
\caption{ \textbf{Visualization of Uncertainty Estimation on the Matterport Dataset}. 
}
 \vspace{5em}
\label{fig:ddp_supp1}
\end{figure*}
\begin{figure*}[hbtp]
\centering
\includegraphics[width=\textwidth]{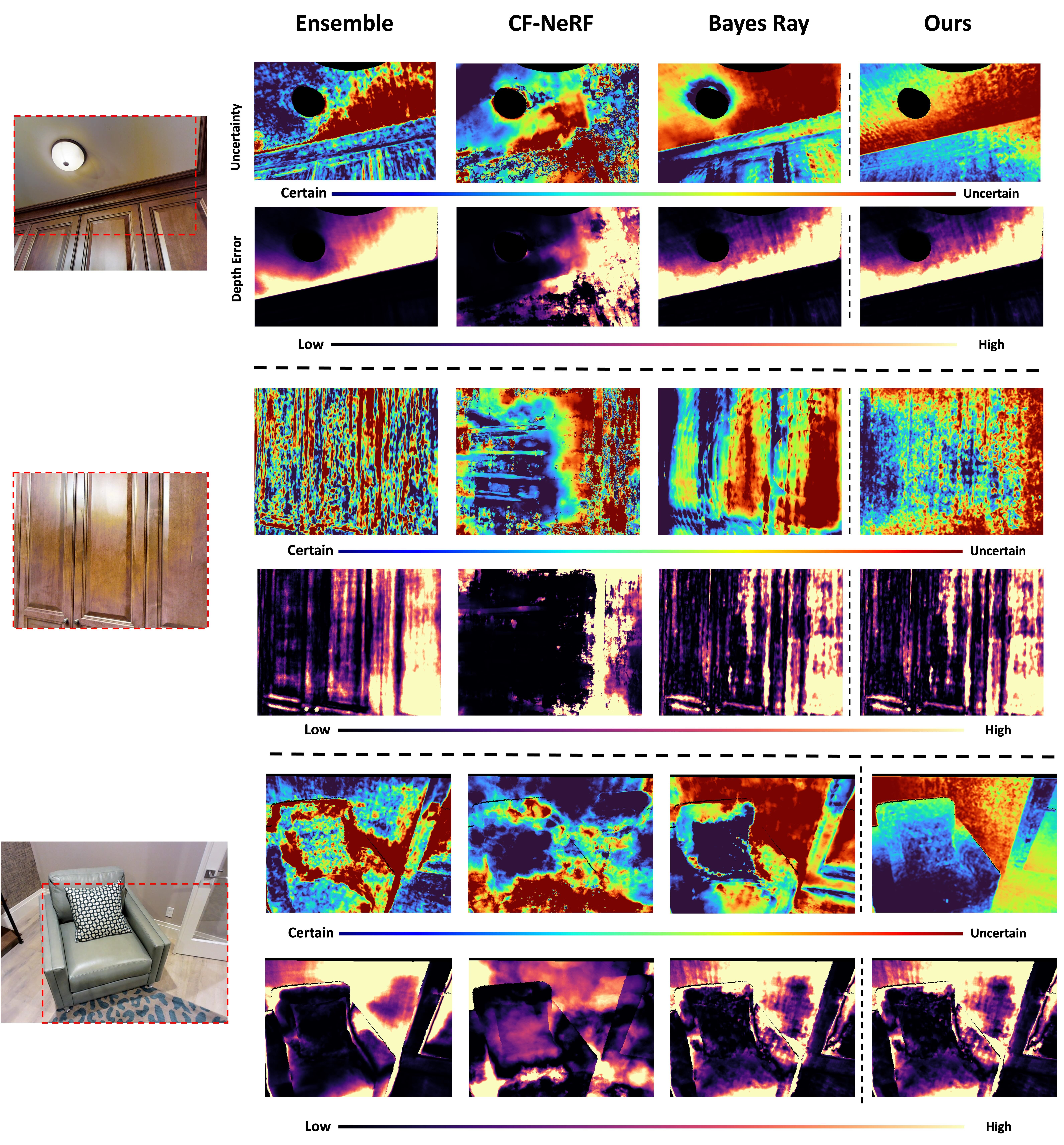}
\caption{ \textbf{Visualization of Uncertainty Estimation on the Matterport Dataset}. 
}
 \vspace{5em}
\label{fig:ddp_supp2}
\end{figure*}

\paragraph{Datasets \& Baselines}
We evaluate uncertainty quantification on two sparse, unconstrained views (outward-facing) datasets -- Scannet~\cite{dai2017scannet} and Matterport3D~\cite{Matterport3D} using the training and test splits released by~\cite{roessle2022depthpriorsnerf}. Each dataset contains three scenes with $17-36$ training and $8$ test images at a room scale. We compare our approach with the previous state-of-the-art CF-NeRF~\cite{CF-NeRF}, concurrent work Bayes' Rays~\cite{bayesrays} and a standard derivation of uncertainty by training an ensemble of NeRFs~\cite{sünderhauf2022densityaware}. We use the same backbone NeRF model as ours when comparing with Bayes' Rays and Ensemble. Following previous works~\cite{s-nerf, CF-NeRF, sünderhauf2022densityaware}, we report the negative log-likelihood (NLL) of $\mathbb{P}(\bm{x}|\{I\})$ to measure uncertainty. We measure the NLL of the ground truth surface under the distributions given each model's uncertainty prediction. Specifically, for Ensemble, we run 10 identical models using different seeds and compute NLL of the ground truth depth under the empirical Gaussian distribution given by the 10 different depth renderings of each test view. For Bayes' Ray, we obtain the posterior likelihood by evaluating their Laplace approximated distribution $p\paren{\bm{\theta}|I}$ at $\theta = 0$. This is equivalent to evaluating the likelihood of $0$ under $\mathcal{N}\paren{\bm{0}, \Diag\paren{\sigma_x, \sigma_y, \sigma_z}}$ for each 3d location $\bm{x} = \paren{x, y, z}$. For a fair comparison, we set Bayes' Ray's prior distribution, which is a zero-centered Gaussian, to have a standard deviation that is half of the diagonal length of the scene bound given by the training camera poses. Lastly for CF-NeRF, we follow their default setting and evaluate the ground truth depth under the empirical Gaussian given by 32 sample rays. 

\paragraph{Additional Visualizations}
Figures~\ref{fig:ddp_supp1} and~\ref{fig:ddp_supp2} demonstrate qualitative comparisons of baselines' and our uncertainty estimation on test images of the Matterport Dataset. Notice that in the last example of Figure~\ref{fig:ddp_supp1}, CF-NeRF's uncertainty does not respect the geometry of the scene, while Bayes' Ray and Ensemble mark the bed as uncertain despite it having low depth error. On the other hand, our method is able to correctly mark the bed as certain thanks to our derived uncertainty estimation based on classical photogrammetry. Moreover, even in challenging settings such as the ceiling of the first example in Figure~\ref{fig:ddp_supp2}, and the closeup view of the closet in the second example of Figure~\ref{fig:ddp_supp2}, our method is able to mark the regions with bad triangulation correctly (i.e., edge of the ceiling) using our formulation. On the other hand, approximated methods such as Bayes' Ray struggle to capture these small uncertainty variations, as Bayes' Ray simply marked the entire ceiling as uncertain. 

Lastly, Figure~\ref{fig:scannet_supp} shows additional visualizations of uncertainty estimations on the Scannet Dataset. Notice that compared to baselines, our method's uncertainty estimation 
is able to achieve better correlations with the depth error map (e.g., The carpet under the desk in the last example, and the tabletop in the second example). 

\paragraph{Implementation Details}
To compute for the posterior likelihood shown in Eq. 13 in the main. Assuming that $\bm{x}$ is uniformly distributed within the scene bound $\Omega$, and that the noise in 2D measurements are independent of each other given $\bm{x}$, we rederive and simplify Eq. 13 as
\begin{equation}
\label{eq:unc2}
    \begin{split}
        & \mathbb{P}\paren{\bm{x} | x_1, \dots, x_K}\\
        &= \frac{\mathbb{P}\paren{x_1, \dots, x_K | 
        \bm{x}}\mathbb{P}\paren{\bm{x}}}{\mathbb{P}\paren{x_1, \dots, x_K}}\\
        & = \frac{\mathbb{P}\paren{x_1, \dots, x_K | 
        \bm{x}}\mathbb{P}\paren{\bm{x}}}{\E_{\bm{x}^\prime \sim \mathbb{P}}\bracket{\mathbb{P}\paren{x_1, \dots, x_K | \bm{x}^\prime}}}\\
        \shortintertext{$\text{Since } x_j's \text{ are pairwise independent given }\bm{x}$,}
        & = \frac{\mathbb{P}\paren{\bm{x}}\prod_{j=1}^K\mathbb{P}\paren{x_j | 
        \bm{x}}}{\E_{\bm{x}^\prime \sim \mathbb{P}}\bracket{\prod_{j=1}^K\mathbb{P}\paren{x_j | 
        \bm{x}^\prime}}}\\
        & = \frac{\mathbb{P}\paren{\bm{x}}\prod_{j=1}^K\mathbb{P}\paren{x_j | 
        \bm{x}}}{\int_{\Omega}\prod_{j=1}^K\mathbb{P}\paren{x_j | 
        \bm{x}^\prime}} \mathbb{P}\paren{\bm{x}^\prime}\,d\bm{x}^\prime\\
        & = \frac{\prod_{j=1}^K\mathbb{P}\paren{x_j | 
        \bm{x}}}{\int_{\Omega}\prod_{j=1}^K\mathbb{P}\paren{x_j | 
        \bm{x}^\prime}}\,d\bm{x}^\prime.
    \end{split}
\end{equation}
The last step is because $\mathbb{P}\paren{\bm{x}} = \mathbb{P}\paren{\bm{x}^\prime} = \frac{1}{\text{Vol}\paren{\Omega}}$. Since the numerator of the final expression in the above derivation is directly computable, we only need to compute the integral in the denominator. Because the computation of $\mathbb{P}\paren{x_j | \bm{x}^\prime}$ involves a non-linear projection operation of $\bm{x}^\prime$ to the image space, the integral is not expressible in analytical form. Thus, we instead use importance sampling over the intersection of the camera frustums to efficiently approximate the integral. 

Specifically, let $\Pi_{j, \bm{x}}$ be the camera frustum given by the camera projection matrix $\text{Proj}_j$ as defined in Sec. 5.1 in the main, and let $\Pi_{j, \bm{x}, \delta}$ be the $\delta$ neighbor camera frustum given by only pixels within $\delta$ range of the center pixel. Then, if we assume that the 2D measurements are corrupted by a zero mean, $\sigma^2$ variance Gaussian noise, it is safe to assume that the probability $\mathbb{P}\paren{x_j| \bm{x}^\prime} \approx 0$ for $\bm{x}^{\prime} \not \in \Pi_{j, \bm{x}, 5\sigma}$ because point outside would fall outside of $5$ standard deviation of the Gaussian, and thereby has negligible probability. Thus, for the product $\prod_{j=1}^K\mathbb{P}\paren{x_j | \bm{x}^\prime}$ to be nonzero, $\bm{x}^\prime$ needs to be in the intersection of all $5\delta$ neighbor frustums $\Pi_{j, \bm{x}, 5\sigma}$. Thus, let $\mathbb{Q}\paren{\bm{x}}$ be a proposal distribution that is uniform within the intersection of all $\Pi_{j, \bm{x}, 5\sigma} \forall j = 1, \dots, K$. Then we can rewrite the last expression in Eq.~\ref{eq:unc2} as 
\begin{equation}
    \label{eq:unc3}
    \begin{split}
        & \mathbb{P}\paren{\bm{x} | x_1, \dots, x_K}\\
        & = \frac{\prod_{j=1}^K\mathbb{P}\paren{x_j | \bm{x}}}{\int_{\Omega}\prod_{j=1}^K\mathbb{P}\paren{x_j | \bm{x}^\prime}}\,d\bm{x}^\prime\\
        & = \frac{\prod_{j=1}^K\mathbb{P}\paren{x_j | \bm{x}}}{\int_{\Omega_{5\delta}}\prod_{j=1}^K\mathbb{P}\paren{x_j | \bm{x}^\prime}}\,d\bm{x}^\prime\\
        & = \frac{\prod_{j=1}^K\mathbb{P}\paren{x_j | \bm{x}}}{\text{Vol}\paren{\Pi_{1:K, \bm{x}, 5\delta}}\E_{\bm{x}^\prime \sim \mathbb{Q}}\bracket{\prod_{j=1}^K\mathbb{P}\paren{x_j | \bm{x}^\prime}}}
    \end{split}
\end{equation}
where 
$$
\Pi_{1:K, \bm{x}, 5\delta} = \bigcap_{j=1}^K \Pi_{j, \bm{x}, 5\sigma}.
$$
Notice that $\Pi_{1:K, \bm{x}, 5\delta}$ is nonempty since $\bm{x}\in \Pi_{1:K, \bm{x}, 5\delta}$. The last expression in the final derivation can be efficiently and robustly approximated using Monte Carlo sampling because $\Pi_{1:K, \bm{x}, 5\delta}$ is the high-density region of $\mathbb{P}\paren{x_j | \bm{x}^\prime}$. In practice, we set the noise's standard deviation to $2$ and sample $10^6$ points to compute the denominator in the final expression of Eq.~\ref{eq:unc3} for each $\bm{x}$. We set $K$ to be $16$ but only take the visible provenance samples defined by a predicted visibility of above $0.7$. This means that each location $\bm{x}$ will have a variable number of associated projection cameras. Additionally, to speed up computation, we compute the posterior likelihood in the closed form if there is only one visible provenance sample from $\mathcal{D}_\theta\paren{\bm{x}}$. 

\begin{figure*}[hbtp]
\centering
\includegraphics[width=\textwidth]{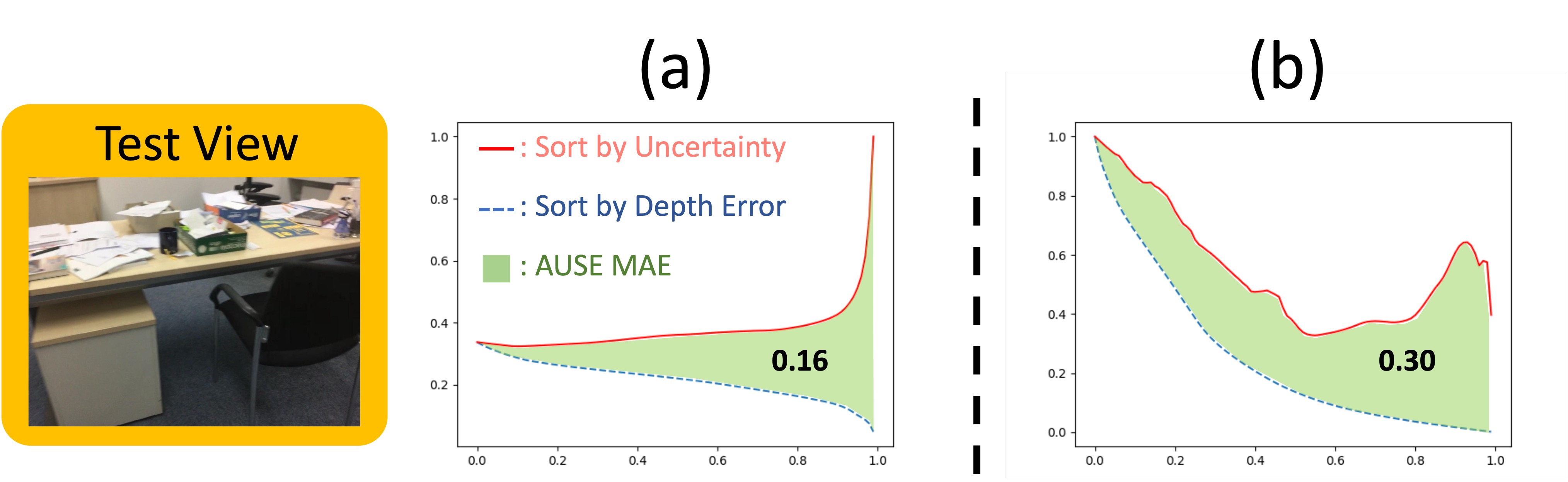}
\caption{ \textbf{Visualization of Sparsification Curves of depth MAE for Bayes' Ray (a) and Ours (b) for the test view on the left}. Notice that despite (a) assigning higher uncertainty to regions with a lower depth error as indicated by its red curve with a positive slope, it incurred a lower AUSE score than method (b)'s curve, which shows a rough correlation w.r.t. depth MAE as shown by its red curve with mostly a negative slope. For visualizations of their uncertainty estimation vis-á-vis depth error, see the last row of Figure~\ref{fig:scannet_supp}.
}
\label{fig:ause}
\end{figure*}

\paragraph{Discussion on AUSE.}
We include a discussion on another metric reported in previous uncertainty estimation literature~\cite{bayesrays} -- Area Under Sparsification Error (AUSE). In our work, we report the negative log likelihood (NLL) following other previous works~\cite{s-nerf, CF-NeRF, sünderhauf2022densityaware} as we found AUSE to be not reflective of the uncertainty of the model in certain scenarios, which will be discussed in this section. AUSE was designed to measure the correlation between the uncertainty estimation and depth error. Concretely to compute for AUSE (see Figure~\ref{fig:ause}), the pixels in each test image are removed gradually (“sparsified”) first i) according to their depth error (blue dotted curve) and then ii) according to their uncertainty estimation (red solid curve). Then the area between the two curves is the the resulting value for the AUSE metric (green region), where lower is considered better. Here, we bring up two discussions on how the AUSE metric may not reflect the methods' uncertainty estimation.

First, the score for AUSE is good if regions with high uncertainty imply that they have high depth error. However, an uncertain region implies that its \emph{variance} is high, which \emph{does not} necessarily imply that the region has large depth error. Such a region may have low depth error but with high variance, which will affect its AUSE score.  


Second, a low AUSE number does not necessarily correspond to a good correlation between a method's uncertainty estimation and depth error. Following~\cite{bayesrays}, Figure~\ref{fig:ause} shows sparsification curves for Bayes' Ray (a) and our method (b) uncertainty estimation w.r.t. to depth mean absolute error (MAE). Notice that, despite (a) obtaining a lower AUSE score (0.16) compared to ours (0.3), the curve sparsified by Bayes' Rays uncertainty has a positive slope throughout the course of sparsification. This behavior implies that for this test view, Bayes' Ray (a) actually assigned higher uncertainty values to regions with a lower depth MAE error. On the other hand, our method's sparsification curve in (b) mostly show negative slopes throughout the course of pixel removal. This indicates that high uncertainty was roughly assigned to regions with higher depth MAE error and vice versa. This observation is consistent with the visualized uncertainty maps in the last row of Figure~\ref{fig:scannet_supp}. Notice that Bayes' Rays ((a) in Figure~\ref{fig:ause} and the third column from the left in Figure~\ref{fig:scannet_supp}) assigned high uncertainty to the drawer and the back of the chair despite these regions having a low depth error as shown in the depth error mean below. On the other hand, our method ((b) in Figure~\ref{fig:ause} and the fourth column from the left in Figure~\ref{fig:scannet_supp}) was able to correctly correlate the uncertainty estimation and the depth error despite it having a higher AUSE score than Bayes' Rays.

\section{Novel View Synthesis}
\label{sec:nvs_imp}
\paragraph{Datasets \& Baselines}
Following previous works~\cite{roessle2022depthpriorsnerf, scade, song2023d}, we evaluate on a subset of ScanNet~\cite{dai2017scannet} released by DDP~\cite{roessle2022depthpriorsnerf} comprising of three scenes each with $17-20$ training and $8$ test views. We also evaluate on a subset of Tanks and Temple released by SCADE~\cite{scade} also comprising of three scenes. For quantitative evaluation, we report the standard metrics PSNR, SSIM~\cite{wang2004image}, LPIPS~\cite{zhang2018unreasonable} on the novel test views. We compare with recent NeRF-based works including state-of-the-art methods DDP~\cite{roessle2022depthpriorsnerf}, SCADE~\cite{scade} and DäRF~\cite{song2023d} on the sparse, unconstrained setting.

\paragraph{Implementation Details}
We initialize our ProvNeRF with pre-trained SCADE~\cite{scade} model and finetune the model with our proposed objective $\mathcal{L}_{\text{ProvNVS}}$ (Eq. 15 main paper) for another $200$K  iterations. The provenance loss is used together with the photometric loss and the space carving loss:
\begin{equation}
     \mathcal{L}_{\methodname} + \mathcal{L}_{\text{ProvNVS}}  = \mathcal{L}_{\text{photometric}} + \beta_1 \mathcal{L}_{\text{space carving}} + \beta_2\mathcal{L}_{\text{ProvNVS}} + \beta_3\mathcal{L}_{\text{fIMLE}}.
\end{equation}
We set $\beta_1 = 0.005$ for all scenes and $\beta_2 = 0.005$ for all scenes except for $\beta_2 = 0.01$ for scene 758 in the Scannet Dataset. $\beta_3 = 0.1$ is set for all scenes. We set the learning rate to be $5e-5$ for all scenes and we only enforce the provenance loss for points with rendered transmittance. 
\input{tables/ablation_nvs_full}
\begin{figure*}
    \centering
    \includegraphics[width=\textwidth]{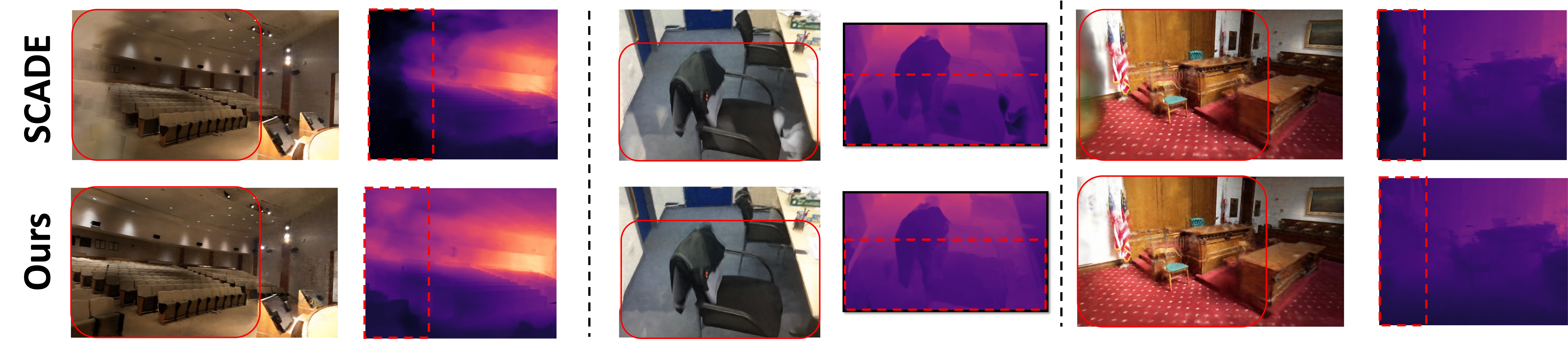}
    \caption{\textbf{NVS Results Shown with Depth Maps.} Compared to the pretrained SCADE, our method additionally improves scene reconstruction by removing clouds in the encircled region of the depth maps.}
    \label{fig:depth_comparison}
\end{figure*}
\paragraph{NVS Depth Comparison}
In Fig.~\ref{fig:depth_comparison} we compare the rendered depth maps before and after applying our NVS regularizer. Notice that our method improves SCADE's depth map by removing floaters in the test views. 
\paragraph{NVS Result Ablation}
We also ablate our choice of fIMLE in the task of novel view synthesis against other probabilistic methods. We compare our approach against the baselines introduced in Section. 5.3 of the main paper in Tab.~\ref{tab:ablation_nvs_full}. Because of our ability to model complex provenance distributions, we achieve better NVS results compared to other probabilistic methods. Notably, jointly training the provenance branch and the underlying NeRF representation further improves the NVS quality, as shown by the comparison of our method with \textbf{Ours}*, which freezes the provenance branch and only updates the underly NeRF weights. The improvement in joint optimization testifies to the mutual benefits between learning provenance and obtaining better geometry. 
\section{Additional Provenance Application: View Selection based on Area Maximization}
\label{viewpoint}
We further demonstrate the usefulness of Provenances for downstream tasks. We show that our per-point provenance can be applied to criteria-based viewpoint selection, which leverages a neural rendering framework to determine the most favorable camera perspective based on a predefined criterion. For instance, this can encompass orienting the camera to align with the normal vector of a specified target or achieving a detailed close-up view of the target. The first criterion can be attained by minimizing the negative dot product between the camera's principal axis and the object's normal, whereas the second can be achieved through the minimization of the negative area of the 2D projection from a 3D plane encasing the object. A naive solution is to directly run gradient descent to optimize each corresponding objective loss $ \mathcal{L}_{\text{obj}} $.

We propose that the integration of two novel viewing loss functions within our model enhances the gradient's informativeness, thereby facilitating an approximation to the objective that is superior to conventional methods. More precisely, for each 3D point $\bm{x}_i$ lying on the object of interest, using our \methodname, we sample provenances $(t_1^{(i)}, \bm{d}_1^{(i)}), \dots, (t_K^{(i)}, \bm{d}_K^{(i)})$ from the distribution $\mathcal{D}_\theta(\bm{x}_i)$. Each distance-direction tuple sample $(t_j^{(i)}, \bm{d}_j^{(i)})$ gives a location $\bm{y}_j^{(i)} = \bm{x}_i - t_j^{(i)}\bm{d}_j^{(i)}$ from which $\bm{x}_i$ is seen, based on this, we define an additional loss $\mathcal{L}_{\text{select}} = -(\mathcal{L}_c + \mathcal{L}_d)$ as follows:
\begin{equation}
\mathcal{L}_c  = \sum_i \max_j \| \bm{y}_j^{(i)} - \bm{c} \|_2^2; \; \mathcal{L}_d = \sum_i \max_j \left( R_{:,3}^\top \bm{d}_j^{(i)} \right),
\end{equation}
where $ [R|\bm{c}] $ denotes the camera pose, $ R_{:,3} $ corresponds to the third column of $R$, reflecting the principal axis of the camera in the global coordinate system. Intuitively, $\mathcal{L}_c$ seeks to align the camera center with the network's prediction, whereas $\mathcal{L}_d$ aims to orient the principal axis of the camera toward the points. By running gradient descent on $\mathcal{L}_{\text{select}} + \mathcal{L}_{\text{obj}}$ from an initial viewpoint of the target object, our method can achieve better PSNR while moving closer to the desired criteria. We benchmark against two baselines: 1) a naive approach that exhaustively searches all training views to locate the one that minimizes $L_{\text{obj}}$, and 2) a direct optimization approach that applies gradient descent solely on $L_{\text{obj}}$ without considering $L_{\text{select}}$.

\begin{figure*}[t]
  \centering
    \includegraphics[width=\textwidth]{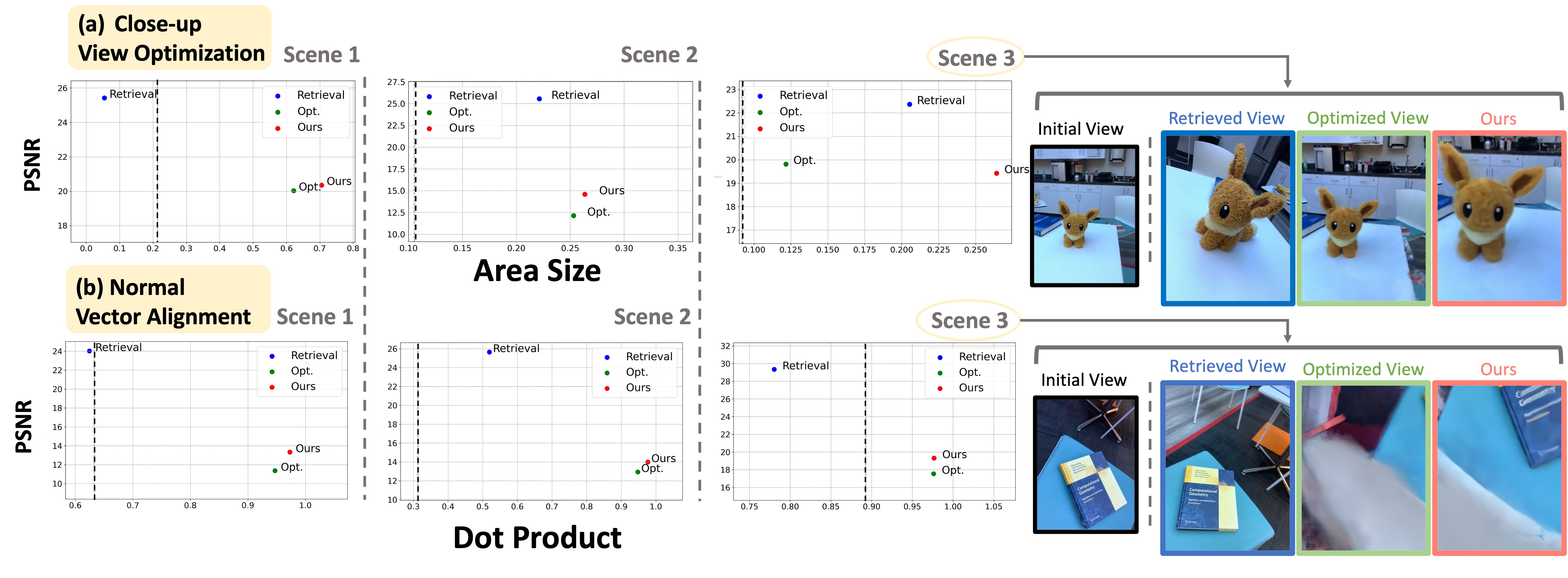}
  \caption{\scriptsize{\textbf{(a)} PSNR and Area Size plots with the objective of maximizing the projected area of the target. \textbf{(b)} PSNR and Dot Product plots the objective of maximizing the dot product between the viewing angle and the target object's normal. The dotted lines denote the objective for the initial views. \textit{(Right)}  We show the final view comparison of our provenance-aided viewpoint selection compared to the baselines under the two objectives in scene 3. Notice that our method (in red) is able to arrive at an objective maximizing view while retaining reconstruction quality. On the other hand, both the retrieval and optimization baselines fail to balance between the two.}}
  \label{fig:area_normal_app}
\end{figure*}


  

\paragraph{Results.}
Both the quantitative (Left) and qualitative (Right) outcomes, presented in Figure~\ref{fig:area_normal_app}, demonstrate that our approach achieves a superior balance between PSNR and the intended objective~\footnote{To compute for PSNR at arbitrary poses, we train an oracle NeRF using the full captured sequence. We use Nerfacto~\cite{nerfstudio} as the oracle NeRF's representation.}. Although the naive retrieval baseline attains a higher PSNR, it fails to satisfactorily meet the desired objective, especially when training views are sparse and limited.
\section{Ablation Study}
\label{sec:ablation_supp}
\paragraph{Metrics Detail}
In the Ablation section~ref{sec_ablation} of main, we computed the Average Precision (AP) and Area Under the Curve (AUC) scores with the confidence score given by the Euclidean distance between the predicted sample and ground truth observation. Here we detail our metric computation. We follow the usual definition of AP and AUC. Given a set of confidence thresholds $\delta_0, \dots, \delta_{N - 1}$, we define
\begin{equation}
\begin{split}
    \text{AP} &= \sum_{k=0}^{N - 1}\bracket{\text{Recall}\paren{k} - \text{Recall}\paren{k + 1}} \cdot \\
    & \text{Precision}\paren{k}
\end{split}
\end{equation}
where $\text{Recall}\paren{N} = 0$ and $\text{Precision}\paren{N} = 1$. On the other hand, AUC is defined as the area under the Precision and Recall curve of the set of thresholds. Thus, to compute for AP and AUC, we need to define how do we measure recall and precision in our setting. To do so, for 3D point $\bm{x}$, we use the set of provenances $\text{Prov}\paren{\bm{x}} \setminus \set{(0, \bm{0})}$ as defined in Eq. 4~\footnote{The visibility term is computed based on the ground truth geometry.} of the main as the ground truth labels, and we sample $K = 128$ provenances from the learned provenance distribution $\mathcal{D}_\theta\paren{\bm{x}}$ and only take the samples with visibility above $0.9$. If we define a discrete uniform distribution over $\widehat{\text{Prov}}(\bm{x})$ as $\mathcal{U}\paren{\bm{x}}$, for a particular confidence threshold $\delta$, the precision and recall at $\bm{x}$ is defined as
\begin{itemize}
\item $\textbf{Precision}\paren{\bm{x}}$ = 
 \begin{equation}
\E_{\paren{\hat{\bm{d}}_i, \hat{t}_i}\sim \mathcal{U}\paren{\bm{x}}}\bracket{\mathbbm{1}_{\set{\min_{j}\norm{\paren{\paren{\bm{d}_j, t_j} - \paren{\hat{\bm{d}}_i, \hat{t}_i}}}_2^2 \, < \, \delta}}}
\end{equation}
 \item $\textbf{Recall}\paren{\bm{x}}$ = 
\end{itemize}
\begin{equation}
    \frac{1}{K}\sum_{i=1}^{K}\mathbbm{1}_{\set{\min_{\widehat{\text{Prov}}(\bm{x})}\norm{\paren{\paren{\bm{d}_j, t_j} - \paren{\hat{\bm{d}}_i, \hat{t}_i}}}_2^2 \, < \, \delta}},
 \end{equation}
We note that \textbf{Precision}$\paren{\bm{x}}$ measures whether all the ground truth observations are covered by the predicted observations, while \textbf{Recall}$\paren{\bm{x}}$ measures if the predicted observations are close to one of the ground truth observations. We use a set of $500$ thresholds log-linearly interpolated from $e^{-20}$ to $1$ for accurate and robust AP and AUC computation. 

\paragraph{Further Ablation Study}
\input{tables/ablation_K}
We further ablate on the number of random function samples $K$ to take during each time of resampling. Tab.~\ref{tab:ablation_k} shows quantitative AP and AUC results on the Scannet dataset. As shown in the table, the quality of the provenance samples improves as we increase the number of latent random function samples. This is because a larger number of sample pools can help the deep transformation better transform the distribution from latent space to the space of provenances. We note that the improvement gradually becomes marginal as $K$ increases beyond $8$. Thus, we use $K=16$ for all of our experiments.  
\paragraph{Implementation of Deterministic Field}
Deterministic Field (see entry ``Deterministic Field" in Tab. 3 main paper) is implemented using the same architecture as the provenance branch of \methodname (i.e., 3 layers of MLP). Input the deterministic field is set to the positionally encoded 3D location and the output is a direction and distance tuple, similarly parameterized as \methodname (For detail of the parameterization, please refer to Sec.~\ref{sec:my_detail}). The deterministic field is trained with the mean square distance of the predicted provenance samples at $\bm{x}$ with an empirical sample from $\hat{\mathcal{D}}\paren{\bm{x}}$. We train the deterministic field for $200$K iterations using the same hyperparameters as our method.

\paragraph{Implementation of Gaussian-Based Models}
The Gaussian-based models (see Gaussian-based w/ C=2, C=5 entries in Tab. 3 main paper) are parameterized using a 3-layer MLP that takes in a 3D location and outputs the means, variances, and weights of each Gaussian. The Gaussian models are trained with the negative log-likelihood of the empirical provenance samples from $\hat{\mathcal{D}}\paren{\bm{x}}$ under the parameterized Gaussian mixture. The models are trained for $200$K iterations with the same set of hyperparameters as \methodname.


\section{Derivation of Objective}
\label{sec_supp_fimle}
In Sec. 4.3 of the main paper, we showed that the fIMLE objective in Eq. 8 of the main paper is equivalent to a pointwise matching loss using the theory of calculus of variation and Euler-Lagrange Equation. We outline the details here.

Specifically, assuming $\bm{D}^{(j)}_{\theta}$ is differentiable for all $j = 1, \dots, K$, we show that any minimum for the L2 norm difference in Eq. 9 of the main paper is only achieved when $\bm{D}_i$ and $\bm{D}^{(j)}_{\theta}$ are pointwise equivalent. Fix $M$ empirical function samples $\hat{\bm{D}}_1, \dots, \hat{\bm{D}}_M$ from $\hat{\mathcal{D}}$, and $K$ i.i.d. sample functions $\bm{D}^{(1)}_\theta, \dots, \bm{D}^K_\theta$ from the distribution $\mathcal{D}_\theta$. Then, we define a functional $J$ of $K$ variables in $\bm{D}^{(1)}_\theta, \dots, \bm{D}^K_\theta$ in the form of
\begin{equation}
\label{eq:el}
    J\paren{\bm{D}^{(1)}_{\theta}, \dots, \bm{D}^{(K)}_{\theta}} = \frac12\sum_{i=1}^M\norm{\hat{\bm{D}}_i - \bm{D}^{(j_i)}_{\theta}}^2_{L^2}
\end{equation}
where 
$$
j_i = \arg\min_{j=1, \dots, K}\norm{\bm{D}_i - \bm{D}^{(j)}_{\theta}}^2_{L^2}
$$ for each $i =1, \dots, M$ denote the argmin of $\norm{\bm{D}_i - \bm{D}^{(j)}_{\theta}}^2_{L^2}$ over $j \in \set{1, \dots, K}$. Then, by the Euler-Lagrange Equation, we know that if the tuple $\paren{\bm{D}^{(1)}_{\theta^*}, \dots, \bm{D}^{(K)}_{\theta^*}}$ is a minimizer to $J$, they must satisfy
\begin{equation}
        \frac{\partial}{\partial \bm{D}^{(j)}_{\theta}}J\paren{\bm{D}^{(1)}_{\theta^*}\paren{\bm{x}}, \dots, \bm{D}^{(K)}_{\theta^*}\paren{\bm{x}}} = 0
\end{equation}
for all $\bm{x}\in \R^3$ and $j = 1, \dots, K$. On the other hand, taking the functional derivative with respect to each of the $K$ variables of $J$, we derive that at point $\bm{x}\in \R^3$,
\begin{equation}
\label{eq:deriv}
\begin{split}
    &\frac{\partial}{\partial \bm{D}^{(k)}_{\theta}}J\paren{\bm{D}^{(1)}_{\theta}\paren{\bm{x}}, \dots, \bm{D}^{(K)}_{\theta}\paren{\bm{x}}}\\
    & = 
    \begin{cases}
        0 & \text{if }k \not\in \set{j_1, \dots, j_M}\\
        \norm{\hat{\bm{D}}_i\paren{\bm{x}} - \bm{D}^{(k)}_{\theta}\paren{\bm{x}}}_2 & \text{if }k = j_i
    \end{cases}.
\end{split}
\end{equation}
Then, combining Eqs~\ref{eq:el} and \ref{eq:deriv}, we see that objective 8 in the main only can achieve its minimum when 
\begin{equation}
\norm{\paren{\hat{\bm{D}}_i - \bm{D}^{(j_i)}_{\theta}}(\bm{x})}_2 = 0,  \;\forall \bm{x}\in \R^3.    
\end{equation}
The above derivation shows that minimizing the $L^2$ differences of functions is equivalent to minimizing the norm differences of the functions at all points. This shows the equivalence of the fIMLE objective in Eq. 8 and \methodname objective Eq. 10 in the main.

\section{Societal Impact}
\label{sec:impact}
Our models require the usage of GPUs both in training time and inference time. We acknowledge that this contributes to climate change which is an important societal issue. Despite this, we observe the improvement in the results that is theoretically grounded and addresses a gap in the existing literature on modeling the locations of likely visibility of NeRFs. To alleviate this issue, we only test our method on selected scenes before scaling up to different scenes across datasets to minimize its impact on climate change.

%% file: tables/ablation_nvs_full.tex
\begin{table}[ht]
    \begin{tabular}{lccc}
    \toprule
         & \small{PSNR} $\paren{\uparrow}$ & \small{SSIM} $\paren{\uparrow}$ & \small{LPIPS} $\paren{\downarrow}$\\
         \midrule
        \small{Determinisic Field}  & 21.38 & 0.720 & 0.307 \\
        Gaussian-based w/ $C = 2$ & 20.77 & 0.702 &	0.336\\
        Gaussian-based w/ $C = 5$& 21.04 & 0.714 & 0.326\\
        VAE-based & 19.28 &	0.657 &	0.334\\\hline
        \small{Frustum Check} & 21.56 & 0.728 & 0.297\\\hline
        \methodname w/ Spatial Inv. $\mathcal{Z}$& 21.63 & 0.731 & 0.294\\
        \textbf{Ours*} & 21.68	& 0.732 &	\textbf{0.291} \\
        \textbf{Ours} & \textbf{21.73} & \textbf{0.733} & \textbf{0.291} \\
        \bottomrule
    \end{tabular}
    \caption{\textbf{NVS Ablation Results on Scannet.} \textbf{Ours}* only updates the underlying NeRF representation without updating the provenance branch.}
    \label{tab:ablation_nvs_full}
\end{table}

%% file: tables/ablation_K.tex
\begin{table}[ht]
    \centering
    \begin{tabular}{lcc}
    \toprule
         K & AP $\paren{\uparrow}$ & AUC $\paren{\uparrow}$ \\
         \midrule
         1 & 0.197 &0.199 \\
        2 & 0.534 & 0.536\\
        4& 0.644 & 0.646\\
        8 & 0.738 & 0.740\\
        16& \textbf{0.745}	& \textbf{0.747}\\
        \bottomrule
    \end{tabular}
    \caption{\textbf{Ablation Results on Scannet.}}
    \label{tab:ablation_k}
\end{table}